\documentclass[twoside]{article}
\usepackage{ecj,palatino,epsfig,latexsym,natbib}

\usepackage{csquotes}

\usepackage{booktabs,tabu} 
\usepackage{multirow}
\usepackage{caption}
\usepackage{algorithm,tabularx}
\usepackage[noend]{algpseudocode}
\usepackage{boldline} 
\usepackage{xcolor}
\usepackage{graphicx}
\usepackage{amsmath}
\usepackage[caption=false,font=normalsize,labelfont=sf,textfont=sf]{subfig}

\makeatletter
\newcommand{\multiline}[1]{%
	\begin{tabularx}{\dimexpr\linewidth-\ALG@thistlm}[t]{@{}X@{}}
		#1
	\end{tabularx}
}
\makeatother

\begin{document}

\ecjHeader{x}{x}{xxx-xxx}{2020}{A Layered Learning Approach for LCSs}{I M Alvarez et al.}
\title{\bf A Layered Learning Approach to Scaling in Learning Classifier Systems for Boolean Problems}

\author{
\name{\bf Isidro~M.~Alvarez} \hfill \addr{yummyhumans@gmail.com}
\AND
\name{\bf Trung B. Nguyen} \hfill \addr{trung.nguyen@ecs.vuw.ac.nz}
\AND
\name{\bf Will N. Browne} \hfill \addr{will.browne@vuw.ac.nz}
\AND
\name{\bf Mengjie Zhang} \hfill \addr{mengjie.zhang@ecs.vuw.ac.nz} 
\AND
\addr{School of Engineering and Computer Science, Victoria University of Wellington, Kelburn, Wellington 6140, New Zealand}}


%


\maketitle


\begin{abstract}

Learning classifier systems (LCSs) originated from cognitive-science research but migrated such that LCS became powerful classification techniques. Modern LCSs can be used to extract building blocks of knowledge to solve more difficult problems in the same or a related domain. Recent works on LCSs showed that the knowledge reuse through the adoption of Code Fragments, GP-like tree-based programs, into LCSs could provide advances in scaling. However, since solving hard problems often requires constructing high-level building blocks, which also results in an intractable search space, a limit of scaling will eventually be reached. Inspired by human problem-solving abilities, XCSCF* can reuse learned knowledge and learned functionality to scale to complex problems by transferring them from simpler problems using layered learning. However, this method was unrefined and suited to only the Multiplexer problem domain. In this paper, we propose improvements to XCSCF* to enable it to be robust across multiple problem domains. This is demonstrated on the benchmarks Multiplexer, Carry-one, Majority-on, and Even-parity domains. The required base axioms necessary for learning are proposed, methods for transfer learning in LCSs developed and learning recast as a decomposition into a series of subordinate problems. Results show that from a conventional tabula rasa, with only a vague notion of what subordinate problems might be relevant, it is possible to capture the general logic behind the tested domains, so the advanced system is capable of solving any individual n-bit Multiplexer, n-bit Carry-one, n-bit Majority-on, or n-bit Even-parity problem.
\end{abstract}


\section{Introduction}


Learning Classifier Systems (LCSs) were first introduced by \cite{holland_adaptation_1975} as cognitive systems designed to evolve a set of rules. LCSs were inspired by the principles of stimulus-response in cognitive psychology \citep{holland_adaptation_1975,holland_adaptation_1976,schaffer1985learning} for interaction with environments. LCSs morphed from being platforms to study cognition to become powerful classification techniques \citep{bull2015brief,butz_rule-based_2006,lanzi_roadmap_1999}. 

An important strength of LCSs is their capability to subdivide the problem into niches that can be solved efficiently. This is made possible by integrating generality into the rules produced. This pressure towards generality means that one classifier could be a solution to a bigger set of problem instances. In the proposed work, we seek to go beyond what the Michigan-style LCS currently offers with its niching strength. We start with XCS, an accuracy-based LCS, which creates accurate building blocks of knowledge for an experienced niche to develop a system to scale to problems of any size in a domain \citep{wilson_classifier_1995,butz2002algorithmic}. 



Although LCS techniques have facilitated progress in the field of machine learning, they had a fundamental weakness. Each time a solution is produced for a given problem, the techniques tend to \enquote*{jettison} any learned knowledge and must start from a blank slate when tasked with a new problem. 

The field of Developmental Learning in cognitive systems contains an idea known as the Threshold Concept \citep{falkner2013computer}. This idea conveys the fact that in human learning there exist certain pieces of knowledge that are transformative in advocating the learning of a task.  These concepts need to be learned in a particular order, thus providing the learner with viable progress towards learning more difficult ideas at a faster pace than otherwise. For instance, humans are taught mathematics in a certain progression; arithmetic is taught before trigonometry, and these two are taught before calculus. The empirical evidence indicates that this sequence will be more effective in fostering the learning of progressively more difficult mathematics~\citep{falkner2013computer}. 


Related to the benefits of the threshold concept are Layered Learning (LL) and Transfer Learning (TL) in artificial systems. In LL, a sequence of knowledge is learned \citep{stone2000layered}. LL requires crafting a series of problems, which enables the learning agent to learn successively harder problems. The benefits of TL are actualised when learning from one domain is transferred to aid learning knowledge to a similar or related domain  \footnote{Some fields define TL as transferring the underlying model \citep{pan_survey_2010} }.  In essence, TL aims to extract the knowledge from one or more source tasks and apply the knowledge to the target task \citep{feng2015memes}.  

Current LCSs can be utilised to extract building blocks of knowledge in the form of GP-like trees, called Code Fragments (CFs). TL can then reuse these building blocks to solve more difficult problems in the same or a related domain. The past work showed that the reuse of knowledge through the integration of CFs into XCS, as a framework, can provide dividends in scaling \citep{iqbal2014reusing}. 

Numerous systems using CFs have been developed. XCSCFC is a system that has extended XCS by replacing the condition of the classifiers with a number of CFs \citep{iqbal2014reusing}. Although XCSCFC exhibits better scalability than XCS, eventually, a computational limit in scalability will be reached \citep{iqbal2013extending}. The reason for this is that multiple CFs can be used at the terminals, as the problem increases in size, then any depth of tree could be created. Instead of using CFs in rule conditions, XCSCFA integrates CFs in the action part of classifiers \citep{iqbal_evolving_2013}. This method produced optimal populations in both discrete domain problems and continuous domain problems. However, XCSCFA lacked scaling to very large problems, even where they had repeated patterns in the data.

In the preliminary work, XCSCF* \citep{alvarez2016human} has applied the threshold concepts, LL, and TL to enable it to solve the n-bit Multiplexer problem. However, this was only a single domain, so the question remains: was the approach robust and easy to implement across multiple domains? Furthermore, the system output was human interpretable after two days' work, where it is needed to generate more transparent solutions to n-bit problems. It is also important to discover ontologies of functions that will map to numerous, heterogeneous patterns in data at scale. This will aid in evolving a compact and optimal set of classifiers at each of the proposed steps \citep{price2005functional}. This work requires hand-crafted layers, where it is usual for humans to specify the problems for learning systems.

In this paper, we aim to develop improvements to XCSCF* that enables it to solve more general Boolean problems using LL. The idea behind this system is still to learn progressively more complex problems using hand-crafted layers. For each tested problem domain, i.e. the Multiplexer, Carry-one, Majority-on, and Even-parity domains, we propose a series of subproblems to enable the LL system to evolve the complex logic behind the tested problems. The Multiplexer and Carry-one problems are ones that lend themselves for research because they are difficult, highly non-linear and have epistasis. In the Multiplexer domain, the importance of the data bits is dependent on the address bits, while, in the Carry-one domain, the first bits of the two half bitstrings occur more frequently with larger niches in the search space. The Majority-on domain is known for its highly overlapped niches, which tend to overwrite optimal rules with over-general ones. Lastly, the Even-parity domain usually requires complex combinations of input attributes to generalise.

The specific research objectives are as follows:

\begin{itemize}
\item \textbf{Develop} methods such that learned knowledge and learned functionality can be reused for Transfer Learning of Threshold Concepts through LL.
\item \textbf{Determine} the necessary axioms of knowledge, functions and skills needed for any system  from which to commence learning.
\item \textbf{Demonstrate} the efficacy of the introduced methods in complex domains, i.e. the Multiplexer, Carry-one, Majority-on, and Even-parity domains. 
\end{itemize}

It is hypothesised that crafting solutions at low-scale problems that scale to any problem in a domain is more plausible and practical than tackling each individually large-scale problem. This is considered a necessary step towards continuous learning systems, which will transition from interrogatable Boolean systems to practical real-world classification tasks.

\section{Background}

\begin{figure}
	\centering
	\includegraphics[width=3.2in]{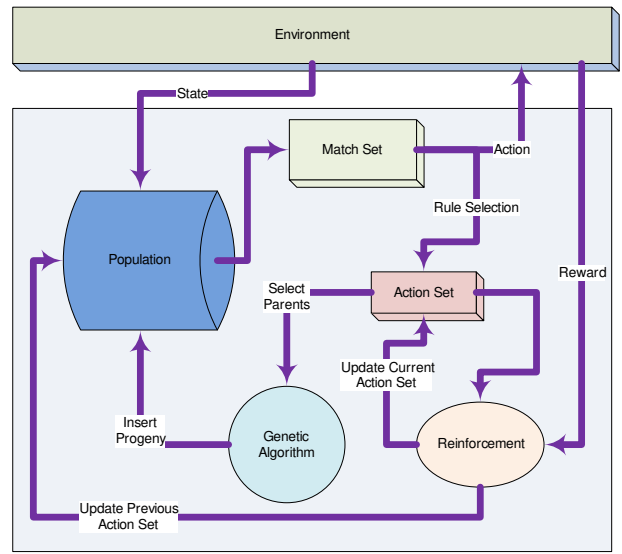}
	\caption{XCS framework showing the processes in the main loop.}
	\label{fig:xcs_framework_overview}
\end{figure}

Figure \ref{fig:xcs_framework_overview} depicts the main highlights of XCS, a Michigan-style LCS developed by Wilson \citep{wilson_classifier_1995}. On receiving a problem instance, a.k.a. an environment state, a match set $[M]$ of classifiers that match the state is created from the rule population. Each available action from the match set is assigned a possible payoff. Based on this array of predicted payoffs, an action is chosen. The chosen action is used to form an action set $[A]$ from the match set. The system executes the chosen action and receives a corresponding reward $\rho$. The action set is updated regarding the reward and the Genetic Algorithm (GA) may be applied \citep{wilson_classifier_1995,butz2002algorithmic}. Subsumption takes place before the offspring are added to the population. If the new population size exceeds the limit, classifiers are chosen to be deleted until the population size is within the valid size. 

XCS differs from its predecessors in a number of key ways: (1) XCS uses the prediction accuracy to estimate rule fitness, which promotes a solution encompassing a full map of the problem via accurate and optimally general rules; (2) evolutionary operations operate within niches instead of the whole population; and (3) unlike the traditional LCS, XCS has no message list and therefore it is suitable for learning Markov environments only \citep{butz2002algorithmic,wilson_classifier_1995}.

Using Reinforcement Learning (RL), XCS guides the population of classifiers towards increasing usefulness via numerous parameters, e.g. fitness.  The main uses of RL are of this mechanism are: 1) identify classifiers that are useful in obtaining future rewards; 2) encourage the discovery of better rules \citep{urbanowicz2009learning}. RL acts independently to covering, where in case there is an empty match set, new rules are created to match the new situation \citep{bull2015brief}. The rules or classifiers are composed of two main parts, the condition and the action. Originally the condition part utilised a ternary alphabet composed of: \{0, 1, \#\} and the action part utilised the binary alphabet \{0, 1\} \citep{urbanowicz_introduction_2017}.



\subsection{CF-based XCSs}
LCSs can select/deselect features using generality through the ``don't care" operator. Originally the don't care symbol was a \enquote*{\#} hash-mark, which comprised part of the ternary alphabet \{0, 1, \#\} \citep{holland_adaptation_1976}.  Since the initial introduction of LCSs, the number of applicable alphabets has been expanded to include more representations such as Messy Genetic Algorithms (mGAs), S-Expressions, Automatically Defined Functions, and Code Fragments. A Code Fragment (CF) is an expression, similar to a tree generated in Genetic Programming \citep{iqbal2012xcsr}.  CFs generate small blocks of code in binary trees with an initial maximum depth of two.  CFs have also been expressed using sub-trees with more than two children, with varying degrees of success \citep{alvarez_reusing_2014}.  The initial depth was chosen, based on empirical evidence, to limit bloating caused by the introduction of large numbers of introns.  Analysis suggests that there is an implicit pressure for parsimony \cite{iqbal2013learningoverlap}.

LCSs based on CFs can reuse learned information to scale to problems beyond the capabilities of non-scaling techniques.  One such technique is XCSCFC. This approach uses CFs to represent each condition bit enabling feature construction in the condition of the classifiers.  The action part uses the binary alphabet \{0, 1\} \citep{iqbal2014reusing}.  An important benefit inherent in CFs is their decoupling between a CF and a position within the condition, i.e. the ordering of the CFs is unimportant. High-level CFs can capture the underlying complex patterns of data, but also pose a large space. The recent LCS, XOF, introduced the Observed List to enable learning useful high-level CFs in rule conditions \citep{nguyen2019online,nguyen2019improvement}. Another way to capture the complex patterns of data is to utilise CFs as rule actions while keeping the ternary representation for rule conditions \citep{iqbal_evolving_2013}.

Previously it has been shown that rule-sets learned by a modified CF-based LCS system, termed XCSCF$^{2}$, can be reused in a manner similar to functions and their parameters in a procedural programming language \citep{alvarez_reusing_in_xcs_2014}. These learned functions then become available to any subsequent tasks. These functions are composed of previously learned rule-sets that map inputs to outputs, which is a straightforward reformatting of the conditions and actions of rules:

\vspace{-3 mm}
\begin{align}
\label{eq:standard}
&'If <Conditions> Then <Actions>' \\
\label{eq:analogous}
&'If <Input> Then <Output>' \\
\label{eq:analogy_function}
&Function(Arguments<Input> Return<Output>) 
\end{align} 
\vspace{-3 mm}

{\parindent0ptEq.} \ref{eq:standard} is the standard way that a classifier would process its conditions to achieve an action, which is analogous to eq. (\ref{eq:analogous}). Eq. \ref{eq:analogy_function} is the analogy of a function.  These functions will take a number of arguments as their input (rule conditions) and will return an output (the effected action of the ruleset) \citep{alvarez_reusing_in_xcs_2014}. 

The technique used in XCSCF$^{2}$ places emphasis on user-specified problems, rather than user-specified instances, which is a subtle but important change in emphasis in EC approaches. That is, the function set is partly formed from past problems rather than preset functions. The advantage of learning functions is that the related CFs (associated building blocks) are also formed and transferred to the new problem, which can bootstrap the search. However, this technique lacks a rich representation at the action part, which will need adapting due to the different types of action values expected in this current work, e.g. binary, integer, and bitstring.


\subsection{Scaling Methods for XCS}
An early attempt at scaling was the S-XCS system that utilizes optimal populations of rules, which are learned in the same manner as classical XCS \citep{ioannides_investigating_2006}. These optimal rules are then imported into S-XCS as messages, thus enable abstraction. The system uses human-constructed functions, such as Multiply, Divide, PowerOf, ValueAt, and AddrOf, among others \citep{ioannides_investigating_2006}. Although these key functions provide the system with the scaffolding to piece together the necessary knowledge blocks, they have an inherent bias and might not be available to the system in large problem domains. For example, in the Boolean domain, the log and multiplication functions do not exist. It also assumes completely accurate populations, whereas the proposed system is required to learn both the population and functionality, from scratch. If supervised learning is permitted (unlike in this work), the heterogeneous approach of ExSTraCS scales well; up to the 135-bit Multiplexer problem \citep{urbanowicz2012instance}. 

%
%
%

Previously, other Boolean problems have been solved successfully by using techniques similar to the proposed work.  One of these is a general solution to the Parity problem described in  \citep{huelsbergen1998finding}.  The technique is similar to the proposed work because it evolves a general solution that is capable of solving parity problems of any length.  It can also address repeating patterns similar to the loop mechanism of the proposed work. On the other hand, this technique makes use of predefined functions making it a top-down approach.  The proposed technique learns new functions, making it more flexible. 

The preliminary work proposed XCSCF* with of various components \citep{alvarez2016human}. Since different types of actions are expected, e.g. Binary, Integer, Real, and String (Bitstring); it is proposed that the functions be created by a system with CFs in the action (XCSCFA), although any rule production system can also be used, e.g. XCS, XCSCFC, etc.  This will facilitate the use of real and integer values for the action as well as enabling it to represent complex functionality.  The proposed solution will  reuse learned functionality at the terminal nodes as well as the root nodes of the CFs since this has been shown to be beneficial for scaling. XCSR would not be helpful here because on a number of the steps, the permitted actions are not a number but a string e.g., kBitString.  Moreover, XCSR with Computed Continuous Action would present unnecessary complications to the work because the conditions of the classifiers do not require real values \citep{iqbal2012xcsr}. Accordingly, it is necessary to explore further ways to expand the preliminary work to adapt to different domains.

\section{The Problems}\label{ss:problems}

In this section, we provide an analytical introduction to the tested problems that enables the training flow in layers. These flows help formalise the intermediate layers in Section \ref{ss:ind_detailed_comp}. The problem understanding also provides an initial guess of the required building blocks (functions and skills) that should be provided beforehand to bootstrap the learning progress of the system. Although even these pre-provided building blocks can also be divided into more elemental knowledge, this work is not to imitate the education of machine intelligence from scratch. Instead, this paper aims to show the ability of XCSCF* to learn progressively more complex tasks, which resemble human intelligence.

One of the underlying reasons for choosing the Boolean problem domains for the proposed work is that humans can solve this kind of problems by naturally combining functions from other related domains along with functionalities from other Boolean problems. Humans are also able to reason that some functions in their \enquote*{experiential toolbox} may be appropriate for solving the problem. The experiential toolbox is the whole of learned functionality for the agent. These functions include \textit{multiplication}, \textit{addition}, \textit{power}, and the notion of a number line. Therefore, the agent here must build-up its toolbox of functions and associated pieces of knowledge (CFs).  A computer program would make use of these functions and potentially many more, but it can not intuit which are appropriate to the problem, and which are not. Therefore, the agent will need guidance in its learning so that it may have enough cross-domain functions to solve the problem successfully. It will need to perform well with more functions than necessary as the exact useful functions may not be known \textit{a priori}. However, at this stage of paradigm development, the agent is not expected to be able to adjust to fewer functions than necessary. The other reason is that Boolean problems are interrogatable so that a solutions to problems at scales beyond enumeration can still be verified.

\subsection{The Multiplexer Domain}
In the Multiplexer problems, the number of address bits is related to the length of the message string and grows along with the length. The search space of the problem is also adequate enough to show the benefits of the proposed work. For example, for the 135-bit Multiplexer the search space consists of $2^{135}$ combinations, which is immensely beyond enumerated search \citep{noauthor_hierarchical_1991}.  

\begin{figure}[ht]
	\centering
	\includegraphics[width=2.8in]{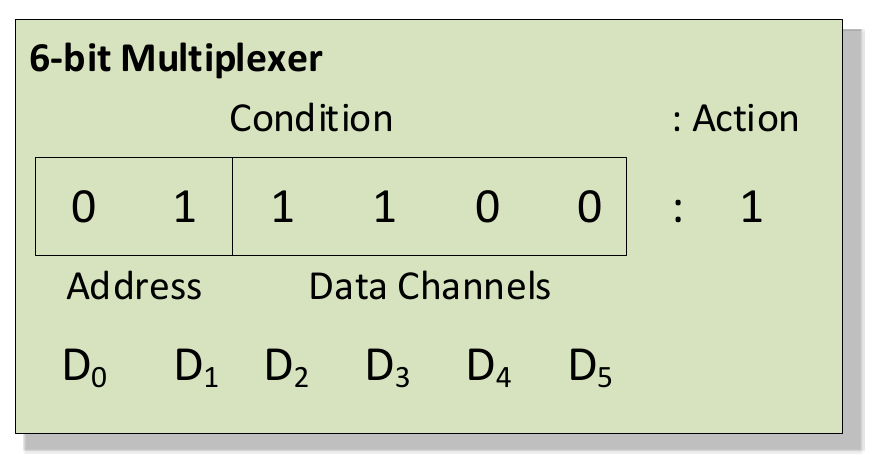}
	\caption{6-bit Multiplexer problem showing the address bits and the data bits of the condition, this distinction is not provided to the learning system.}
	\label{fig:mux_example}
\end{figure}

An example of a 6 bit Multiplexer is depicted in Figure \ref{fig:mux_example}.  Determining the number of address bits \textit{k} requires using the \textit{log} function, as depicted in equation \ref{eq:kbits}, in this example \textit{k} is 2.  Then \textit{k} bits must be extracted from the string of bits to produce the two address bits.  The next step is to convert the address  bits into decimal form; this requires knowledge of the \textit{power base 2} function as well as elementary \textit{looping}, \textit{addition} and \textit{subtraction} functions.  Depending on the approach to this step, \textit{multiplication} may also be required.  The two address bits translate to 1 in decimal form, as shown in figure \ref{fig:mux_example}, i.e. D\textsubscript{0} and D\textsubscript{1}.  The decimal number points to the data bit D1 that contains the value to be returned.  The index begins at 0 and proceeds from the left towards the right, as shown in Figure \ref{fig:mux_example}.

Besides functions, the experiential toolbox will also contain skills.  These are capabilities that the agent will have learned or will have been given beforehand; one example is the looping skill. Skills, unlike functions, do not have a return value, but can manipulate pointers to values (e.g. move around a bitstring).  For example, a human understands all the operations required for counting \textit{k} number of bits, starting from the left of the input string.  Then a human would have to conceptualize how to convert the address bits to decimal, which requires the ability to multiply and add.  If we wanted to increase the difficulty level, we could have the human determine the number of \textit{k} address bits required for a particular problem:

\begin{equation}
\label{eq:kbits}
k = \left\lfloor \log_{2}L \right\rfloor   
\end{equation}   
Equation \ref{eq:kbits} determines the number of \textit{k} address bits by using the length of the input.  In this case the person would need familiarity with the log base 2 function as well as the floor function.  A human would eventually determine the address bits with increasing difficulty but a software system would have to learn this functionality before even attempting  to solve the n-bit Multiplexer problem.

\subsection{The Carry-one Domain}

The Carry-one domain is the set of problems that checks whether the addition of two numbers, in the form of binary numbers, carries one in the addition of the highest-level bits of the two numbers. Binary numbers are represented by bitstrings. The input of Carry-one problems is a bitstring concatenating the two bitstrings of the two binary numbers to be added. 

Humans can approach this problem in various ways. In this work, we design a training flow to check whether the summation of two binary numbers, as a bitstring, has a length higher than the length of the two half-bitstrings representing the two binary numbers. First, the learning agent should learn to detach the two half-bitstrings to obtain the two binary numbers. However, the learning agent has no idea about which parts represent the bitstrings of the two binary numbers. Therefore, the first training step is to teach the learning agent to obtain the half length of the input attributes. Then, the next step is to train the agent to extract two half-bitstrings as the two binary numbers to be added with the knowledge of how many bits to be used. After that, the learning agent is required to obtain the bitstring representing the result of the binary summation. Finally, the last training stage is to check whether a bit $1$ is carried at the highest-level bit. Humans can anticipate that the solutions for these processes would possibly require the following skills and functions: binary addition, head list extraction, tail list extraction, value comparison, division, and constant (the ``Half Length" problem would require a constant number of value $2$).

\subsection{The Even-parity and Majority-on Domains}

Even-parity problems check whether the number of bits $1s$ in the input bitstring is even. The operations of this problem domain are straightforward. We devised the training flow with two steps in Section \ref{ss:ind_detailed_comp}, including ``Sum Modulo 2" and ``Is Even-parity" problems. For Majority-on domain, the learning agent is asked to check whether the majority of bits in the input are $1s$. We can anticipate that this problem domain would involve with the subproblem ``Half Length" from the Carry-one domain. The second training step is to teach the learning agent to compare the summation of bits $1s$ with the output of the ``Half Length" problem. These two domains would be expected to require summation of bitstring, modulo $2$, constant, and comparison skills.

\subsection{Individual Detailed Components}\label{ss:ind_detailed_comp}

\begin{figure}[ht]
	\centering
	\includegraphics[width=3.05in]{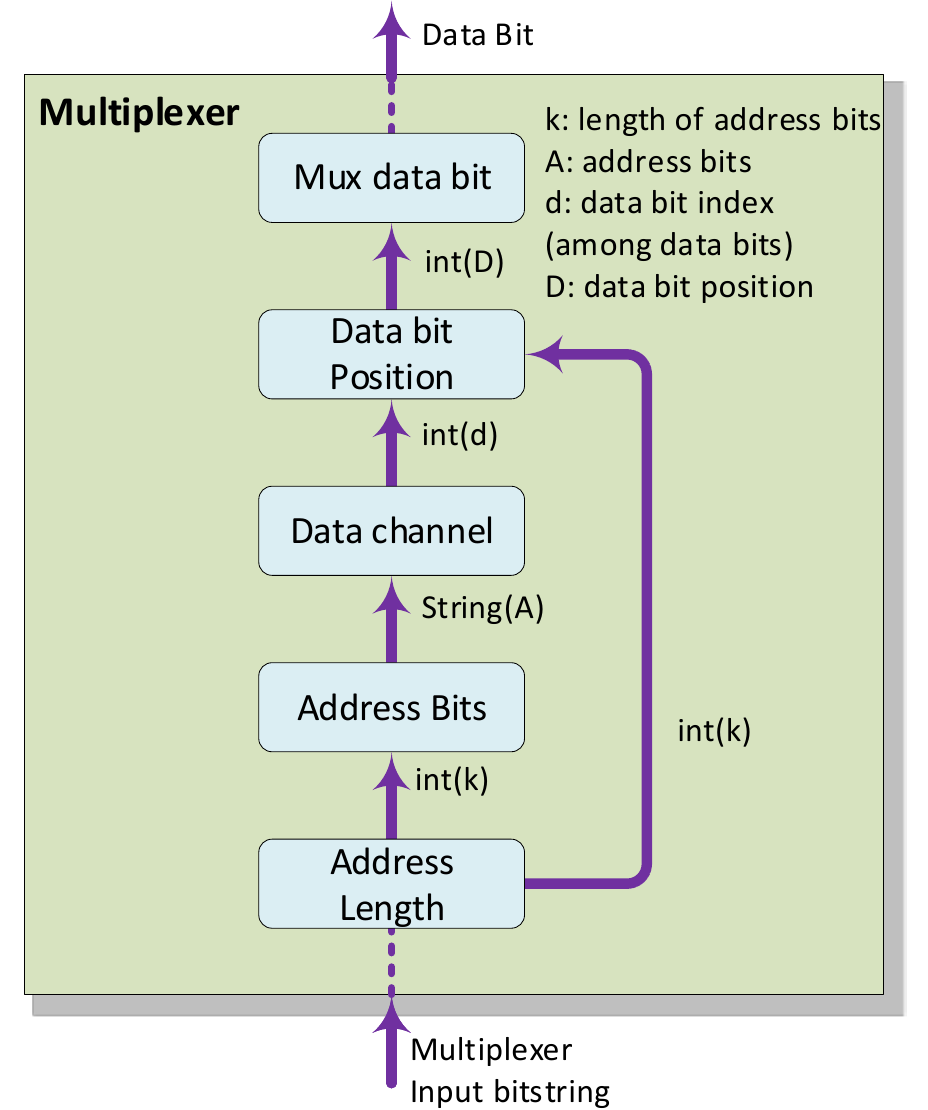}
	\caption{Multiplexer training flow. Each stage of this flow is designed to obtain an aspect of the logic behind the Multiplexer domain. These stages follow the analysis of the Multiplexer problem in Section \ref{ss:problems}.}
	\label{fig:mux_message_functions_inputs}
\end{figure}

\begin{figure}[ht]
	\centering
	\includegraphics[width=3.05in]{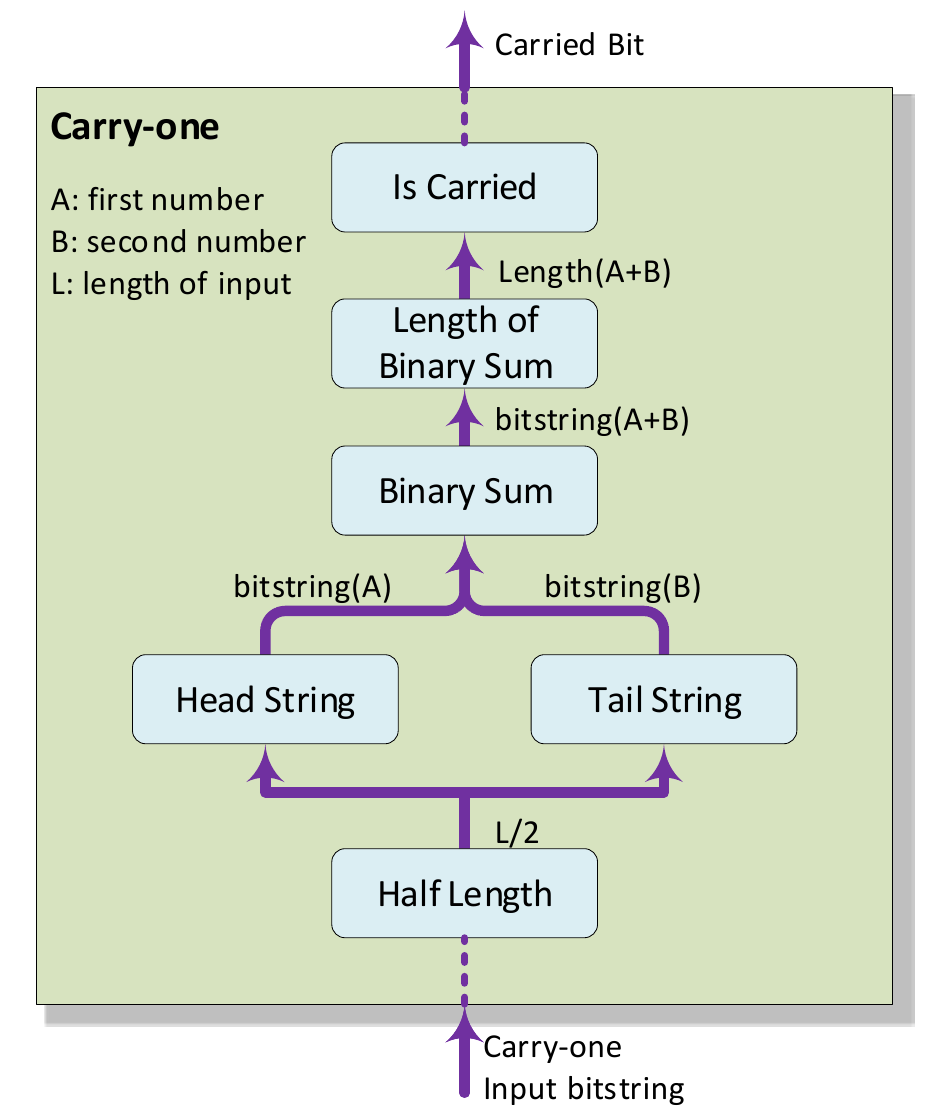}
	\caption{Carry-one training flow.}
	\label{fig:carryone_training}
\end{figure}

\begin{figure}[ht]
	\vspace{-3mm}
	\centering
	\includegraphics[width=4.0in]{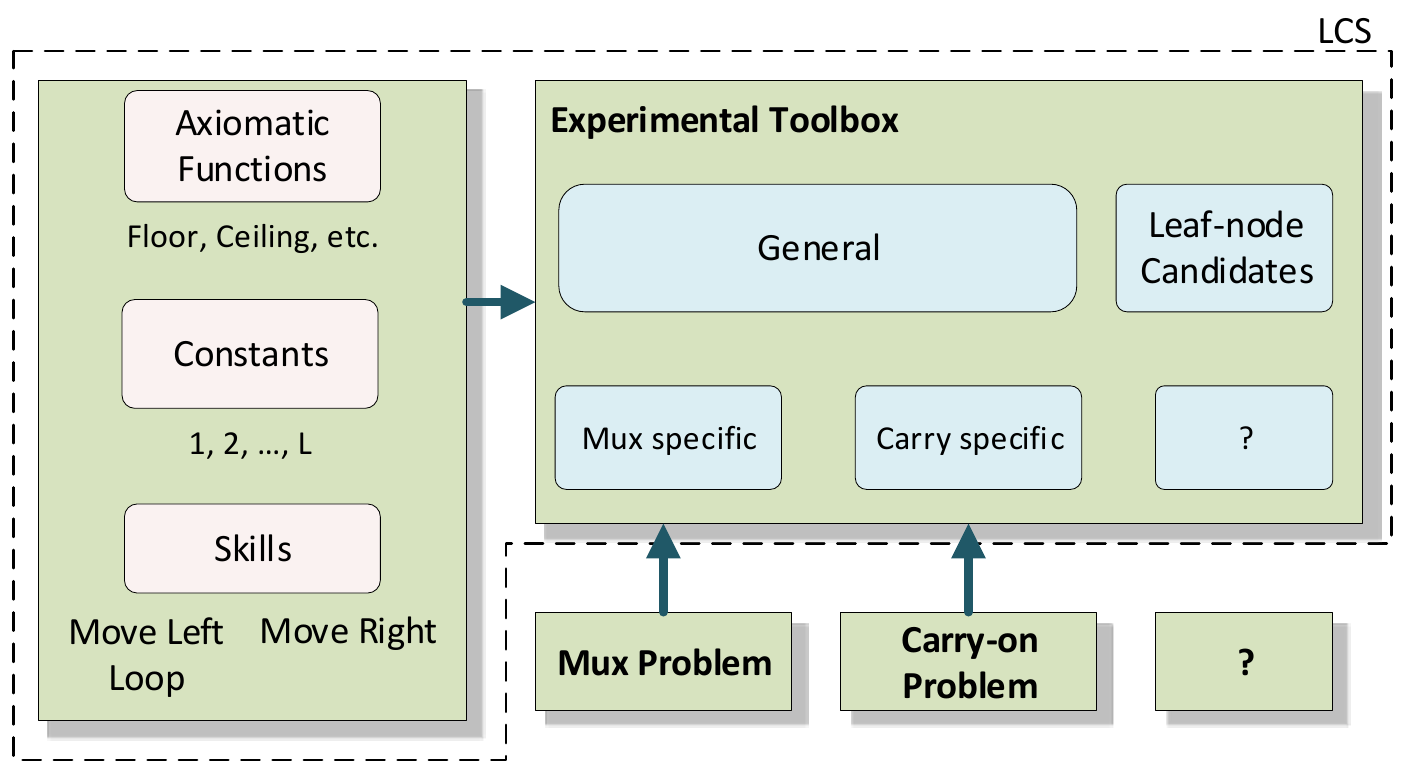}
	\caption{Training encompasses different types of functions, skills and axioms. The experiential toolbox will contain general and problem specific learned functionality. The question marks indicate the next domain and functionality learned from it.}
	\label{fig:method}
	\vspace{-3mm}
\end{figure}

According to the analysis above, the probable methods of separating the Multiplexer and Carry-one domains are shown in Figure \ref{fig:mux_message_functions_inputs} and Figure \ref{fig:carryone_training} respectively. These training flows require a ``human teacher" to form these ``curricula". The training flow of the Multiplexer domain has five main steps corresponding to the first six subproblems listed below, while the Carry-one domain is divided into six other subproblems from ``Half Length" to ``Is Carried" problems. Each subsequent part builds upon the rules learned from the previous step as well as from the Axioms provided. Figure \ref{fig:method} illustrates  the relationships between the Axioms, skills and learned functionality and their CF representation in Multiplexer subproblems.  The figure also depicts how the type of problem faced can feed domain specific functionality into the experiential toolbox of the system.  This is shown by the arrow flowing from the Multiplexer domain towards the Experiential Toolbox. All subproblems for the four benchmark problems are described below with samples provided in the Supplementary material. Table \ref{tab:functions_learned} shows the set of the functions to be learned, note that these were furnished in order, as a curricula. These functions correspond to the subproblems of the curricula.

At each step, the system has access to leaf node candidates, hard-coded functions, the learned CFs, and learned CF-ruleset functions. Table \ref{tab:skills_provided} shows a listing of all the skills made available to the system along with their system tags (used to interpret results) and their input/output data types. This function list is anticipated to be useful based on the above analysis of the benchmark problem domains. In addition to required skills, we also provide extra skills that complement the anticipated ones. These skills could possibly provide unexpected solutions or at least test the ability of XCSCF* to ignore redundant irrelevant skills. 


\begin{table}[!ht]
	\centering
	\caption{Functionality Provided (Hard-coded functions)}
	\begin{tabular}{|l|c|l|l|} \hline
		Functions & Tags & Input & Output\\ \hline
		Floor & $[$ & Float & Integer \\ \hline
		Ceiling & $]$ & Float & Integer \\ \hline
		Log & $\{$  & Float & Float \\ \hline
		Length & $L$ & String & Integer \\ \hline
		Power 2 Loop (binary to decimal) & $2d$ & String & Integer \\ \hline
		Add & $+$ & Floats, Integers & Integer \\ \hline
		Subtract & $-$ & Floats, Integers & Float \\ \hline
		Multiply & $*$ & Floats & Integer \\ \hline
		Divide & $/$ & Floats & Integer \\ \hline
		ValueAt & $@$ & String, Integer & Binary \\ \hline
		Constant & $c$ & None & Integer \\ \hline
		StringSum & $sum$ & String & Integer \\ \hline
		BinaryAddition & $\oplus$ & Strings & String \\ \hline
		BinarySubtraction & $\ominus$ & Strings & String \\ \hline
		HeadList & $($ & String, Integer & String \\ \hline
		TailList & $)$ & String, Integer & String \\ \hline
		isGreater & $>$ & Floats, Integers & Binary \\ \hline
		Modulo & \% & Integers & Integer \\
		\hline\end{tabular}
	\label{tab:skills_provided}
\end{table}

\renewcommand{\arraystretch}{1.1}
\begin{table}[!ht]
	\centering
	\caption{Functions to be learned.}
	\begin{tabular}{|l|c|l|l|} 
		\hline
		Functions & Tags & Input & Output\\ \hlineB{2}
		
		KBitsGivenLength & $k_l$ & Integer & Integer \\ \hline
		KBits & $k$ & String & Integer \\ \hline
		KBitString & k\textsubscript{s} & String & Integer \\ \hline
		Bin2Int & $b2d$ & String & Float \\ \hline
		AddressOf & $d_c$ & String & Float \\ \hline
		ValueAt & $M@$ & String & Binary \\ \hline
		HalfLength & $h$ & Integer & Integer \\ \hline
		HeadString & $S_h$ & String & String \\ \hline
		TailString & $S_t$ & String & String \\ \hline
		BinarySum & $S_+$ & String & String \\ \hline
		LengthBinarySum & $L_+$ & String & Integer \\ \hline 
		isCarried & $iC$ & String & Boolean \\ \hline
		SumMod2 & $sm2$ & String & Integer \\ \hline
		isEvenParity & $iP_e$ & String & Boolean \\ \hline
		isMajorityOn & $iM$ & String & Boolean \\ \hline
	\end{tabular}
	\label{tab:functions_learned}
\end{table}

It is important to note that the work presented here does not seek to provide a learning plan for a system to follow and ultimately arrive at the solution to a given problem. The aim here is to facilitate learning in a series of steps, where in this case the learned functionality could potentially help a system to arrive at a general solution to \textit{any} set problem.  In other words, it is important for the system to learn to mix and match the different learned functions in a way contributive to learning; a way that will produce a general solution. The number of subordinate problems can always be increased in the future, e.g. learning basic functions such as an adder or a multiplier via Boolean functions or even learning the log function via training data.

\subsubsection*{Multiplexer Address Length - kBitsGivenLength}

The first step is to determine the number of \textit{k} address bits that will contribute to the solution for the n-bit Multiplexer.  The length function (Table \ref{tab:skills_provided}) furnishes the system with the length of the environment state instead of the constant $L$ in the previous version of XCSCF* \citep{alvarez2016human}.  The training data-set used consists of instances of possible lengths and the corresponding number of address bits.

\subsubsection*{Multiplexer Address Length - kBits}

This step is to determine the number of \textit{k} address bits when the input is the original input of the Multiplexer problem. The training dataset in this problem replaces the input lengths of the previous ``kBits given Length" problem ($k_l$) with the input bitstring of the Multiplexer problem at various scales.

\subsubsection*{Multiplexer Address Bits - kBitString}

This part extracts the first \textit{k} bits from a given input string.  The data-set will be random bit strings, say length 6, and a given \textit{k} length where the action is the first \textit{k} bits.

\subsubsection*{Multiplexer Data Channel - Bin2Int}

This problem trains the learning agent to convert a binary number to a decimal integer.  This is crucial because the system needs this information to determine the position of the data bit.  However, this is not a trivial task as the system would need to be cognizant of many functions that a human would potentially already have in their experiential toolbox.  The data-set will be random strings with the action being the equivalent integer number.

\subsubsection*{Multiplexer Data Bit Position - AddressOf}

This functionality determines the location of the data bit given the input bitstring. This problem is to guide the learning agent to discover the addition of the address length and the decoded data channel. The data-set will be random strings and decoded address with the integer action.

\subsubsection*{Multiplexer Data Bit - ValueAt}

The functionality to be learned is to return the bit referenced from a bitstring. The system is trained using a dataset of bitstrings of varied lengths (from $3$ to $20$) with a reference integer and corresponding output bit. This problem is actually a Multiplexer problem with variable scales.

\subsubsection*{Sum Modulo 2 - SumMod2}
This problem is the first step of training the Even-parity problem domain. It determines whether the total number of bits $1$ in the input bitstring is even or odd. The ground truth of this problem is the summation of all bits in the input bitstring modulo $2$.

\subsubsection*{Is Even-parity - isEvenParity}

This step is the final step of training the Even-parity domain. This problem expects $True$ if the number of bits $1$ in the input bitstring is even and $False$ otherwise. It is variable in size (from 1-bit to 11-bit Even-parity problems) to encourage only general solutions for the Even-parity problem domain. With varied scales, the solution can solve the problem at any scale. The Even-parity problems with relatively small scales are already intractable to traditional XCS with the ternary encoding because XCS must form a one-to-one mapping of instances to rules.

\subsubsection*{Half Length - HalfLength}
This problem is a regression problem, which requires the learning agent to return the half-length of the input bitstring. This problem can provide prerequisite knowledge for both the Carry-one and Majority-on domains.

\subsubsection*{Head String - HeadString}
This is a step in the training flow of the Carry-one domain, see Figure \ref{fig:carryone_training}. It trains the learning agent to obtain the first number of the addition, which is the first half of the input bitstring. The outputs are binary numbers represented by bitstrings.

\subsubsection*{Tail String - TailString}
This step is similar to the ``Head String" problem but the expected output is the latter half of the input bitstring or the second number of the addition.

\subsubsection*{Binary Summation of Two Strings - BinarySum}
This problem requires the learning agent to add the outputs of the two preceding problems, which are the output of two input numbers in the Carry-one domain and also represented by bitstrings.

\subsubsection*{Length of Binary Sum - SumStringLength}
The expected output of this problem is the length of the output from the preceding problem. This is to learn to predict the length of the binary number resulted by adding the binary number of the first half and the binary number of the second half.

\subsubsection*{Is Carry-one - isCarried}
This requires the general logic behind the Carry-one problem domain. It is to determine whether $1$ is carried at the highest bit when adding the binary number of the first half and the binary number of the second half. The scales of this problem were set to vary from 2-bit to 12-bit.

\subsubsection*{Is Majority On - isMajorityOn}
This problem is the final stage of training the Majority-on domain. It expects a returned value of $True$ if more than half the bits in the input are on ($1$), and $False$ otherwise. The size of input bitstrings are randomly selected from the range of $[1,7]$ bits.

\section{The New Method}

This work disrupts the standard learning paradigms in EC, where the goal is to learn abilities using a top-down approach, by aligning it with LL. The proposed work uses a bottom-up approach by learning functions and using parts or entire functions to solve more difficult problems.  In other words, the method here is to specify the order of problems/domains (together with robust parameter values) while allowing the system to automatically adjust the terminal set through feature construction and selection, and ultimately develop the function set. This is analogous to a school teacher determining the order of threshold concepts for a student in a curricula \citep{meyer2006overcoming}. The system can use learned rule-sets as functions along with the associated building blocks, i.e. CFs, that capture any associated patterns; this is an advantage over pre-specifying functionality.

This method modifies the intrinsic problem from finding an overarching \enquote*{single} solution that covers all instances or features of a problem to finding the structure (links) of sub-problems that construct the overall solution. Learning the underlying patterns that describe the domain is anticipated to be more compact and reusable as they do not grow as the domain scales (unlike individual solutions that can grow impractically large as the problem grows, e.g. DNF solutions to the Multiplexer problem).

We employed an adapted XCSCFA as the algorithm for the agents of XCSCF* that learn subproblems. This XCSCF* has the type-fitting property that can: (1) verify the type compatibility between connected nodes within generated CFs; and (2) the output type of CFs is compatible with the required actions of the current problem environment\footnote{There are sufficient novel contributions to XCSCF* to warrant a new acronym, but as the old one is now superseded and the LCS field already has many acronyms, XCSCF* is retained.}.

\subsection{Type-fitting XCSCFA}\label{ssect:xcscfa2}

In addition to type-fitting CFs, the next important adjustment is pre-provided CFs have been introduced to this version of XCSCF*. Table \ref{tab:leaf_node_candidattes} describes pre-provided CFs, which are all candidates for leaf nodes of CFs. First, leaf node candidates include the CFs representing input attributes, called ``base CFs'', which resemble base CFs in XOFs \citep{nguyen2019online}. 

In the previous version of XCSCF* \citep{alvarez2016human}, a constant $L$ for the length of input bitstrings is provided as a possible leaf node for CFs. This feature is infeasible when the subproblems have variable scales. Therefore, the second change of to XCSCF* to provide another base CF listing all attributes (termed $attlst$) in the order provided by the problem. We hypothesise that this new feature improves the generality of the system by providing inborn knowledge. Lastly, to learn more general problems, it is necessary to provide a system with arbitrary constants. In the limit of Boolean problems, CF generation can access constants (constant CFs) of values from $1$ to the \enquote*{length of the current input attributes} as possible leaf nodes. The constant $L$ in the previous implementation can be obtained by the provided function $Length$ in Table \ref{tab:skills_provided}.

\begin{table}[]
	\vspace{-3 mm}
	\centering
	\caption{Leaf node candidates. $LEN$ is the length of input bitstring.}
	\begin{tabular}{|l|c|l|} \hline
		Leaf node candidates & Tag & Type \\ \hline
		Base CFs of separated attributes & $D_0,D_1,etc.$ & type of attributes \\
		List of all attributes & $attlst$ & String \\
		Constants & $1,2,...,LEN$ & Integer \\ 
		\hline\end{tabular}
	\label{tab:leaf_node_candidattes}
\end{table}

\subsubsection*{Type-fitting Code Fragments}\label{sssect:typed_cfs}

Inspired by Strongly-Typed GP \citep{montana_strongly_1995}, we propose here the type-fitting property for CFs to reduce the search space by fitting each node with only compatible inputs and outputs. CFs with the new type-fitting property, called typed CFs, are designed to create workable and eligible CFs. Being workable refers to the compatibility of the output of CFs with the expected actions of the target problem and the compatibility among the function nodes of a CF. The output type of a function in the node $cf_i$ must be compatible with the input types of the function in the node $cf_{j}$ that takes $cf_i$ as input. Being eligible includes two conditions: the output type of the root node function must be compatible with at least one of the action types of the problem, and the leaf nodes are CFs from the leaf-node candidates. For example, when selecting a leaf node that is the first input to a function that requires the first input to be of type String (e.g. $sum$, $@$, $2d$, etc., see Section \ref{ss:ind_detailed_comp}), the only compatible leaf node candidate is the $attlst$. Ultimately, the type-fitting property keeps learning agents from generating unworkable CFs.

Accordingly, generating typed CFs applies a top-down recursive process of generating tree nodes, i.e. the function \textit{genNode} illustrated in Algorithm \ref{alg:TCFs_generation}. We keep the depth limit of CFs as $2$, as is the original definition of CFs \citep{iqbal2014reusing}. Generating a new CF needs to match with the action types of the problem and available output types from the leaf-node candidates. First, the top node of a typed CF must employ a function with output types compatible with the action types of the problem. Then the process recursively builds lower-level nodes that satisfy the type-fitting property. At any point when generating nodes, there is also a fixed probability of $0.5$ for generating a leaf node from the leaf-node candidates, which stops the CF from going any deeper.

\begin{algorithm}[ht]
	\begin{algorithmic}[1]
		\Procedure{genNode}{$T_o,T_i,l_i$}
		\State $T_i'=\phi$
		\If{$l_i=2$}
		\State Output types $T_o=T_a$
		\EndIf
		\If{$l_i=1$}
		\State Output types $T_{i'}=T_b \cup \{integer\}$
		\EndIf
		\State \multiline{Filter function set $S_{filtered}$ from $S_f$ by required output types $T_o$ and input types $T_i$}
		\State Function $f=randomSelect(S_{filtered})$
		\For{index $i$ in $f.inputs$}
		\If{$l_i-f.level>0$ and $random[0,1)<0.5$}
		\State \multiline{$f.inputs[i]$=\Call{genNode}{$f.input\_types[i]$,$T_{i'},l_i-f.level$}}
		\Else
		\State Set of compatible base CFs $S_{bCF}=\phi$
		\If{integer $\in function.input\_types[i]$}
		\State $c=randomSelect([1,...,len(Atts)])$ 
		\EndIf
		\For{$cf_{base}$ in all base CFs}
		\If{\multiline{$cf_{base}.out\_types \& f.input\_types[i] \neq \phi$}}
		\State Add $cf_{base}$ to $S_{bCF}$
		\EndIf
		\EndFor
		\State $f.inputs[i]=randomSelect(S_{bCF})$
		\EndIf
		\EndFor
		\EndProcedure
	\end{algorithmic}
	\caption{Typed CFs are generated based on a recursive function for generating nodes. The function is given the set of action types $T_a$, the type set of base CFs $T_b$, the expected output types $T_o$, the expected input types $T_i$, the intermediate level $l_i$, and a clustered set of all functions $S_f$.}
	\label{alg:TCFs_generation}
\end{algorithm}

To reduce the search space and generate verifiable CFs, it is necessary to have compatibility rules among the four value types (Binary, Integer, Real, and String). We followed the sense of numerics as well as the type compatibility of the programming language (Python) to devise compatibility rules among types. Boolean variables are compatible with integers and floats, and integers are compatible with floats, the compatibility does not follow the opposite way. Lists are not compatible with other types.

\section{Results}

\subsection{Experimental Setup}

The experiments were executed $30$ times with each having an independent random seed. The stopping criterion was when the agent completed the  number of training instances allocated, which were chosen based on preliminary empirical tests on the convergence of systems.  The proposed systems were compared with XCSCFC and XCS. The settings for the experiments are common to the LCS field \citep{urbanowicz_introduction_2017} and similar to the settings of the previous version of XCSCF* \citep{alvarez2016human}. They were as follows: Payoff $1,000$; the learning rate $\beta = 0.2$; the Probability of applying crossover to an offspring $\chi = 0.8$; the probability of mutation $\mu=0.04$; the probability of using a don't care symbol when covering $P\_{don'tCare} = 0.33$; the experience required for a classifier to be a subsumer $\Theta_{sub} = 20$; the initial fitness value when generating a new classifier $F_{I} = 0.01$; the fraction of classifiers participating in a tournament from an action set $0.4$. In addition, error threshold $\epsilon_0$ was set to $10.0$. This new XCSCF* naively uses the same population size $N=1000$ for all problems.

\subsection{Experimental Tests}

\begin{figure*}[!hbt]
	\centering
	\hspace{-0.40in}	
	\subfloat[kBitsGivenLength]{
		\label{fig:mux_seris:mux_0}		
		\includegraphics[width=1.81in]{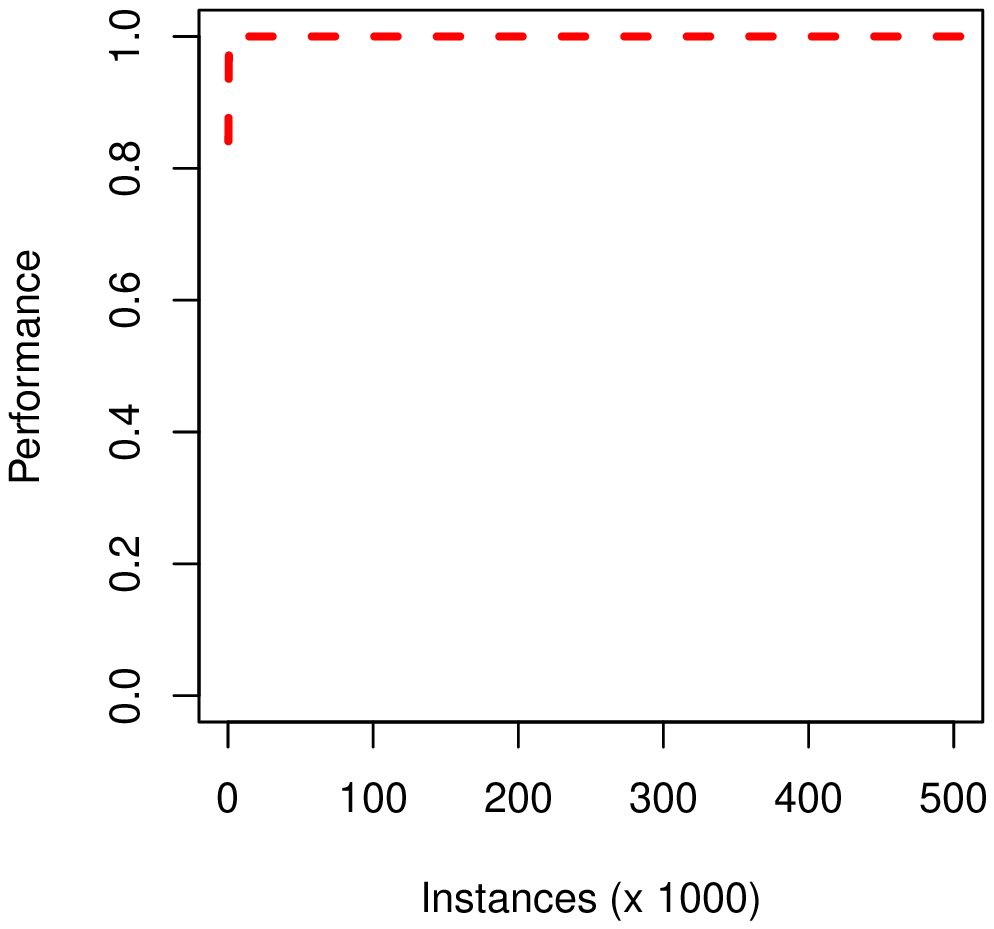}}
	\hspace{-0.01in}	
	\subfloat[kBits]{
		\label{fig:mux_seris:mux_1}		
		\includegraphics[width=1.81in]{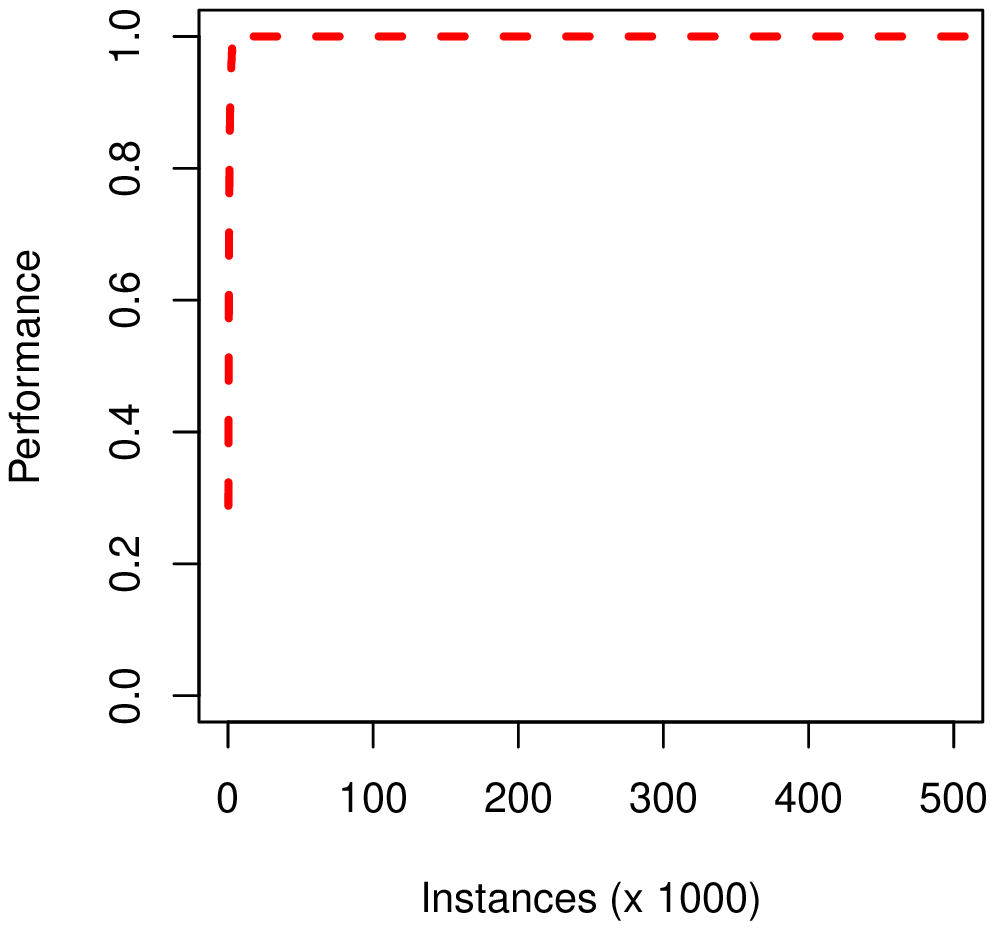}}
	\hspace{-0.01in}	
	\subfloat[kBitString]{
		\label{fig:mux_seris:mux_2}		
		\includegraphics[width=1.81in]{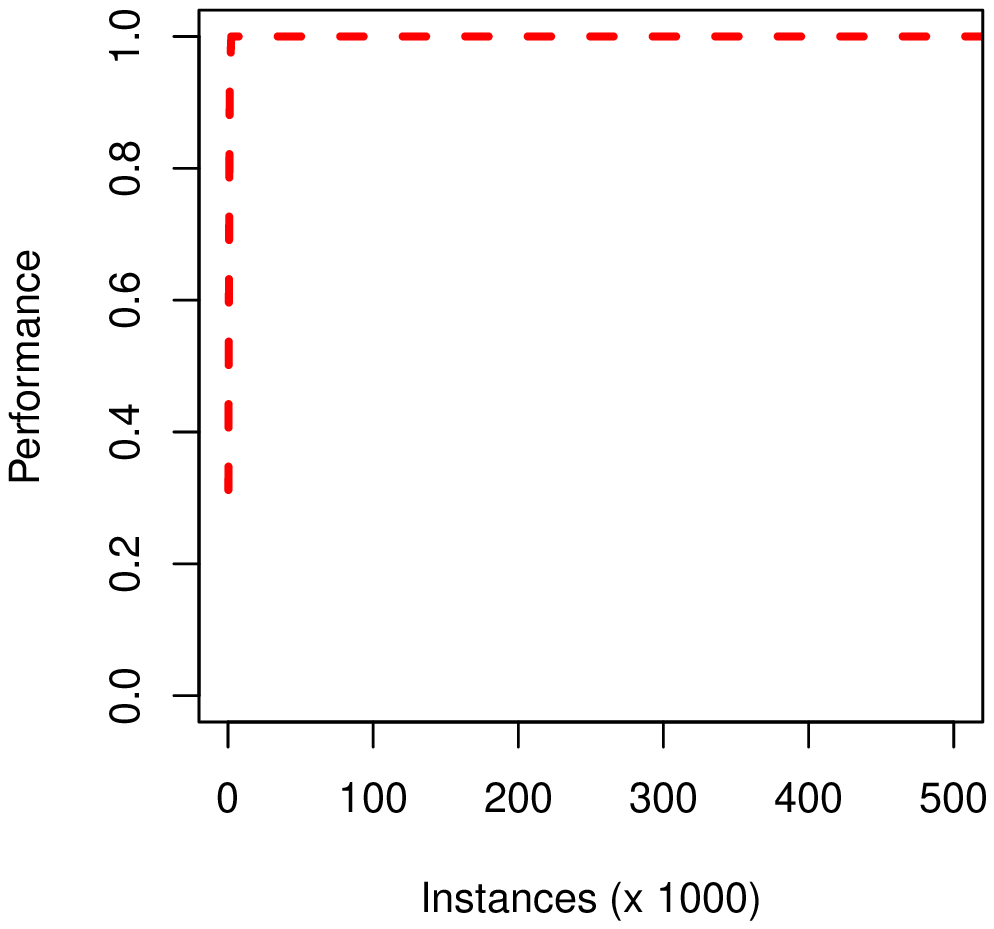}}
	\\
	\hspace{-0.4in}	
	\subfloat[Bin2Int]{
		\label{fig:mux_seris:mux_3}		
		\includegraphics[width=1.81in]{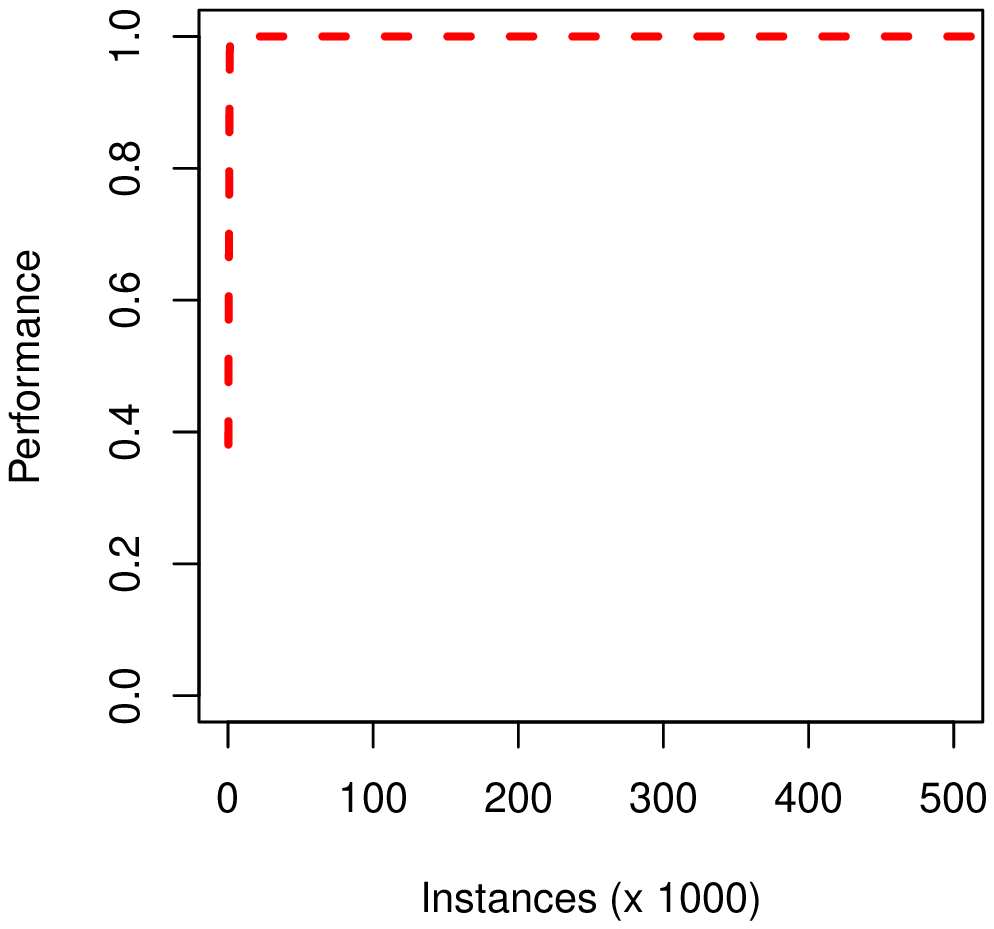}}
	\hspace{-0.01in}	
	\subfloat[AddressOf]{
		\label{fig:mux_seris:mux_4}		
		\includegraphics[width=1.81in]{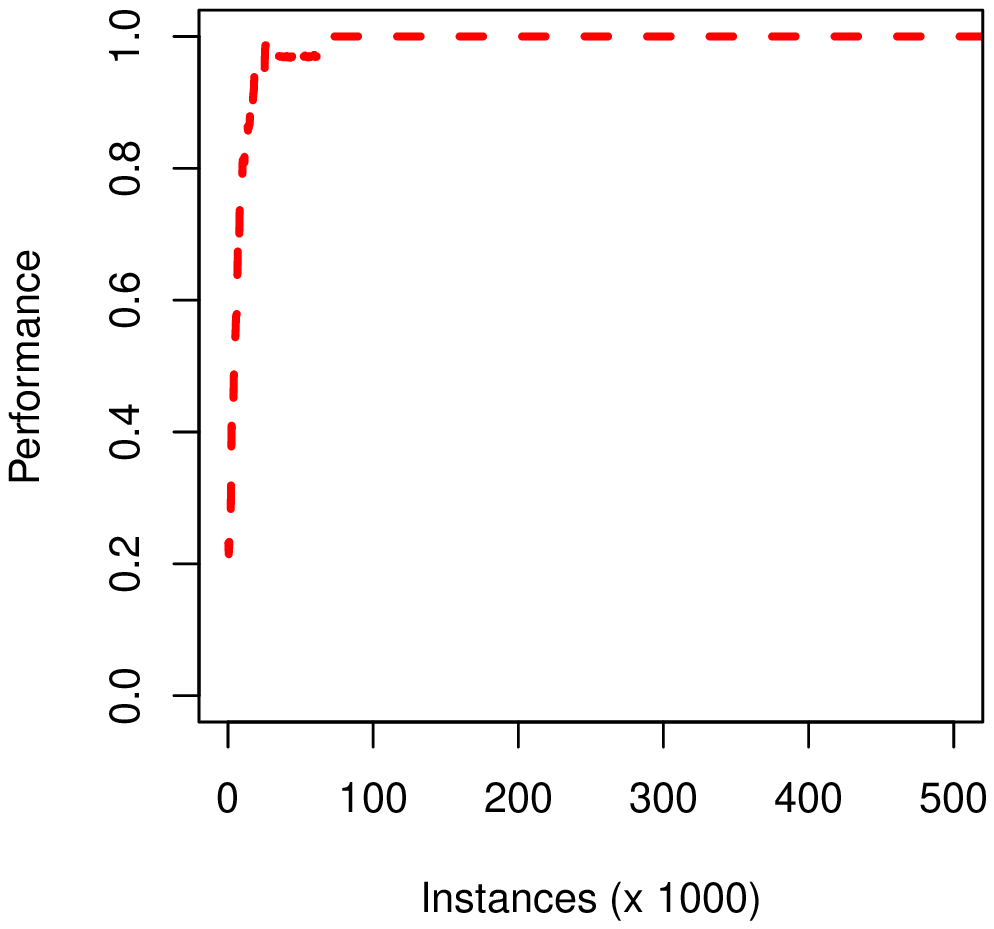}}
	\hspace{-0.01in}	
	\subfloat[ValueAt]{
		\label{fig:mux_seris:mux_5}		
		\includegraphics[width=1.81in]{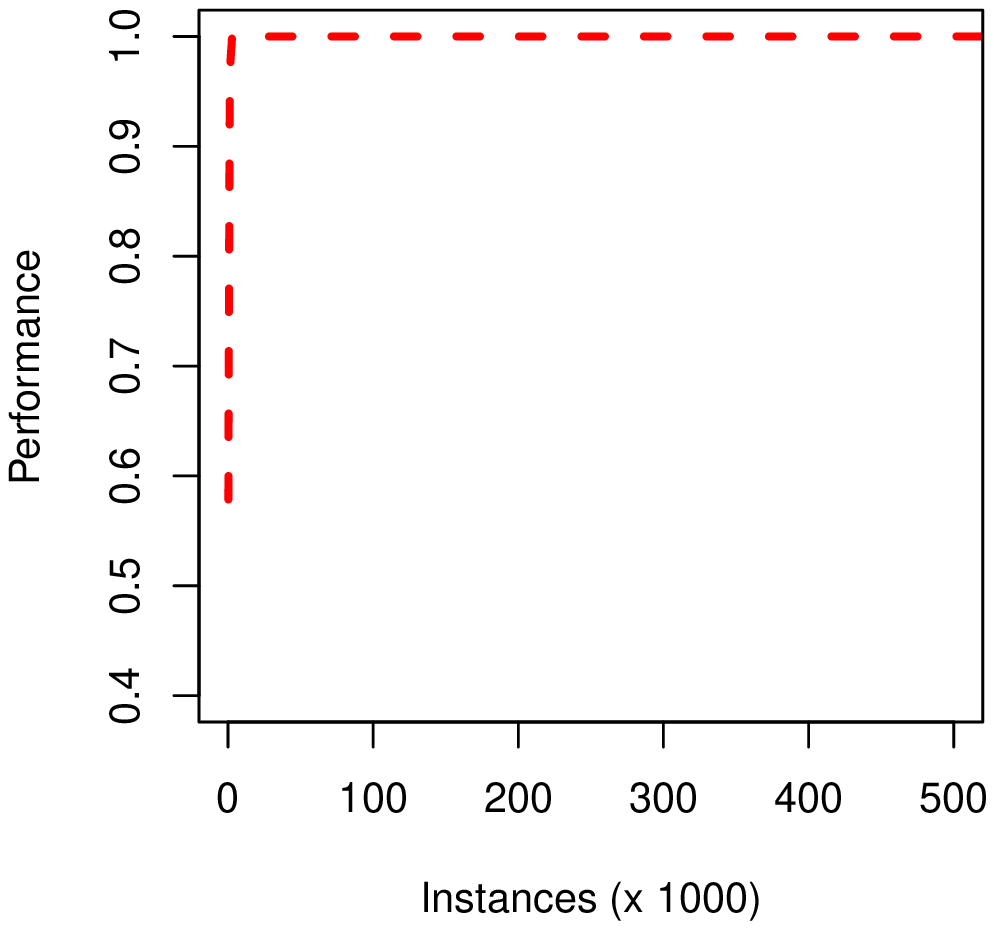}}	
	\captionsetup{justification=centering}
	\caption{Learning curves of the subproblems of the Multiplexer domain.}
	\label{fig:mux_seris}				
\end{figure*}
\begin{figure*}[hbt]
	\centering
	\hspace{-0.40in}	
	\subfloat[HalfLength]{
		\label{fig:car_series:car_0}		
		\includegraphics[width=1.81in]{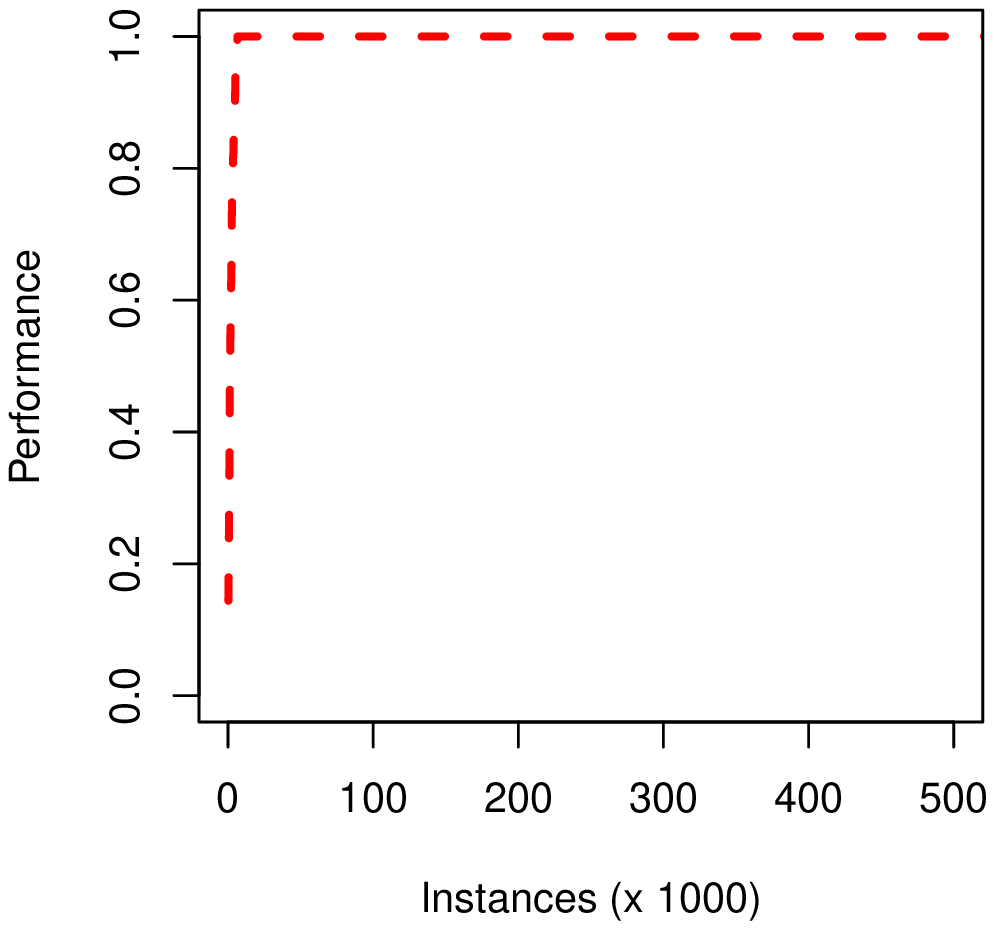}}
	\hspace{-0.01in}	
	\subfloat[HeadString]{
		\label{fig:car_series:car_1}		
		\includegraphics[width=1.81in]{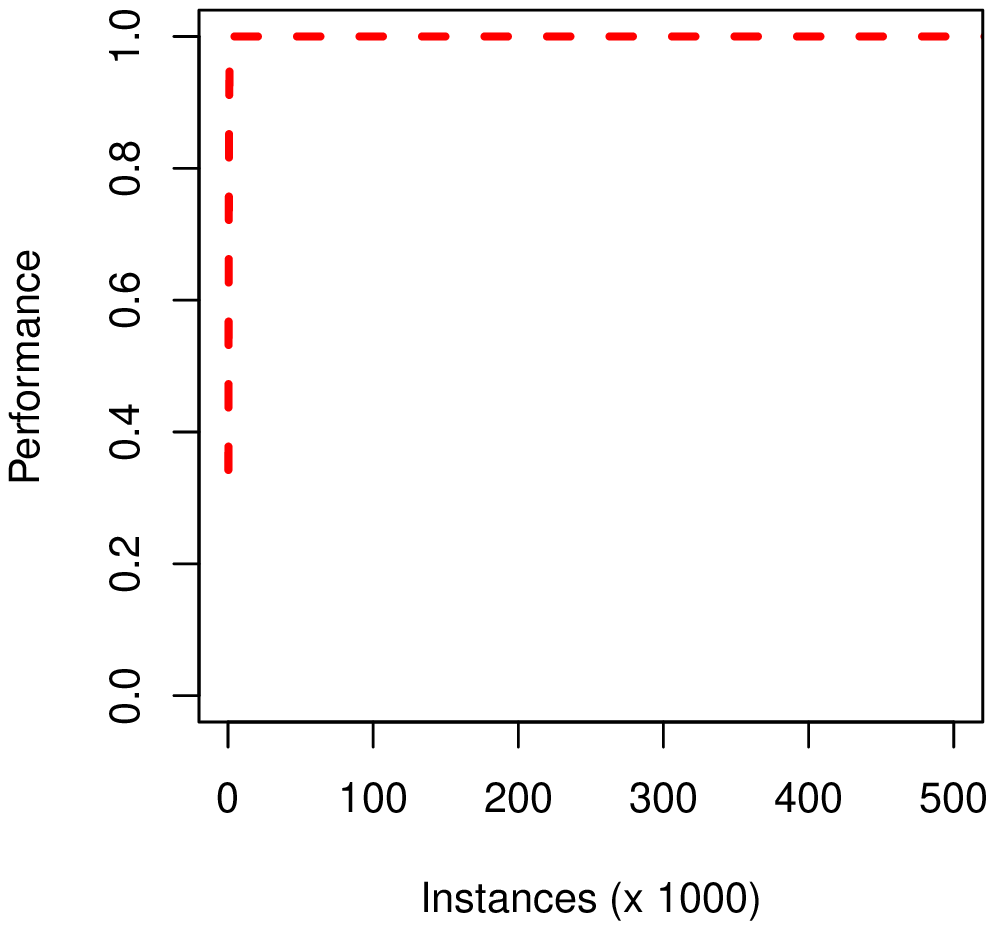}}
	\hspace{-0.01in}
	\subfloat[TailString]{
		\label{fig:car_series:car_2}		
		\includegraphics[width=1.81in]{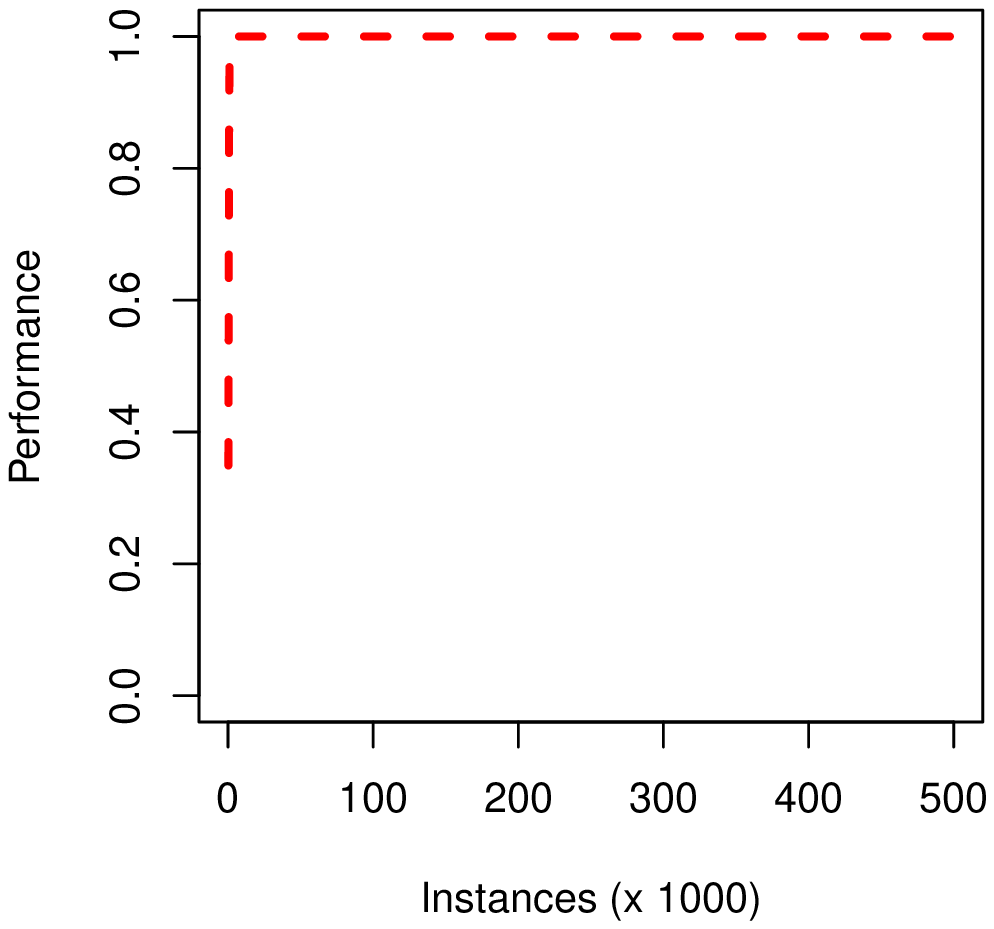}}
	\\
	\hspace{-0.40in}	
	\subfloat[BinarySum]{
		\label{fig:car_series:car_3}		
		\includegraphics[width=1.81in]{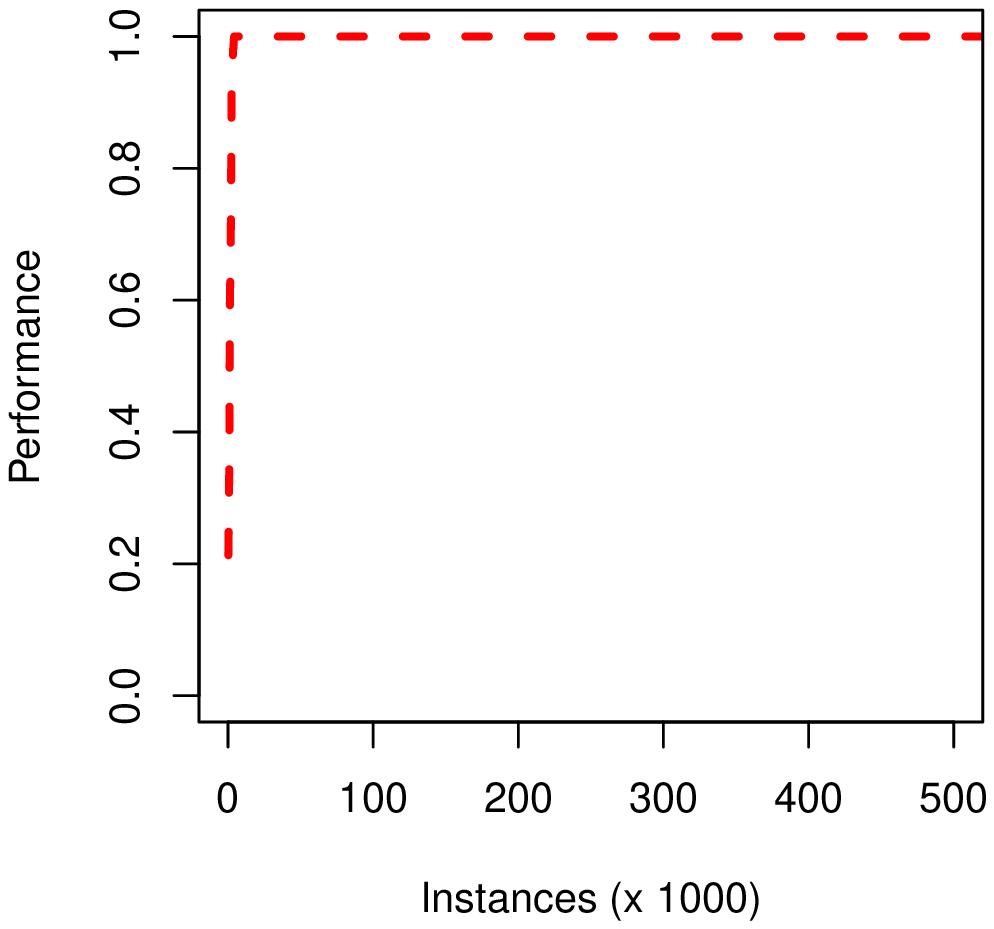}}	
	\hspace{-0.01in}
	\subfloat[SumStringLength]{
		\label{fig:car_series:car_4}		
		\includegraphics[width=1.81in]{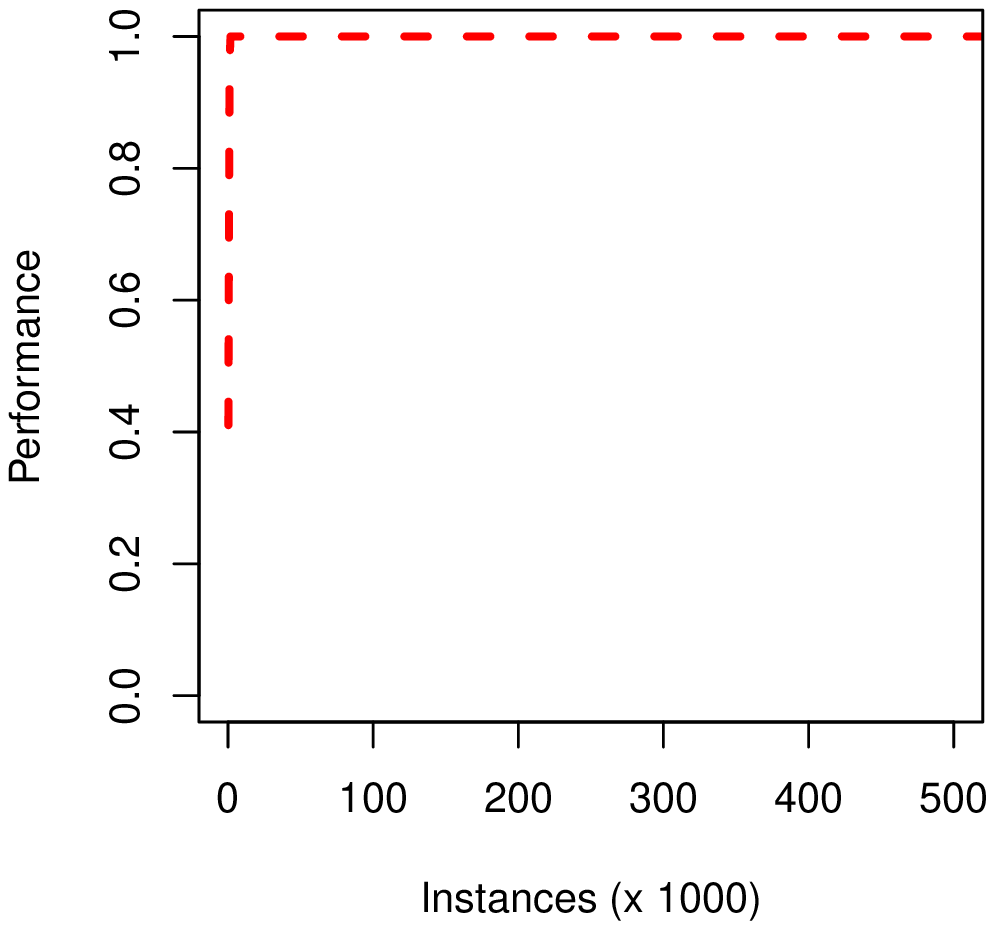}}	
	\hspace{-0.01in}
	\subfloat[isCarried]{
		\label{fig:car_series:car_5}		
		\includegraphics[width=1.81in]{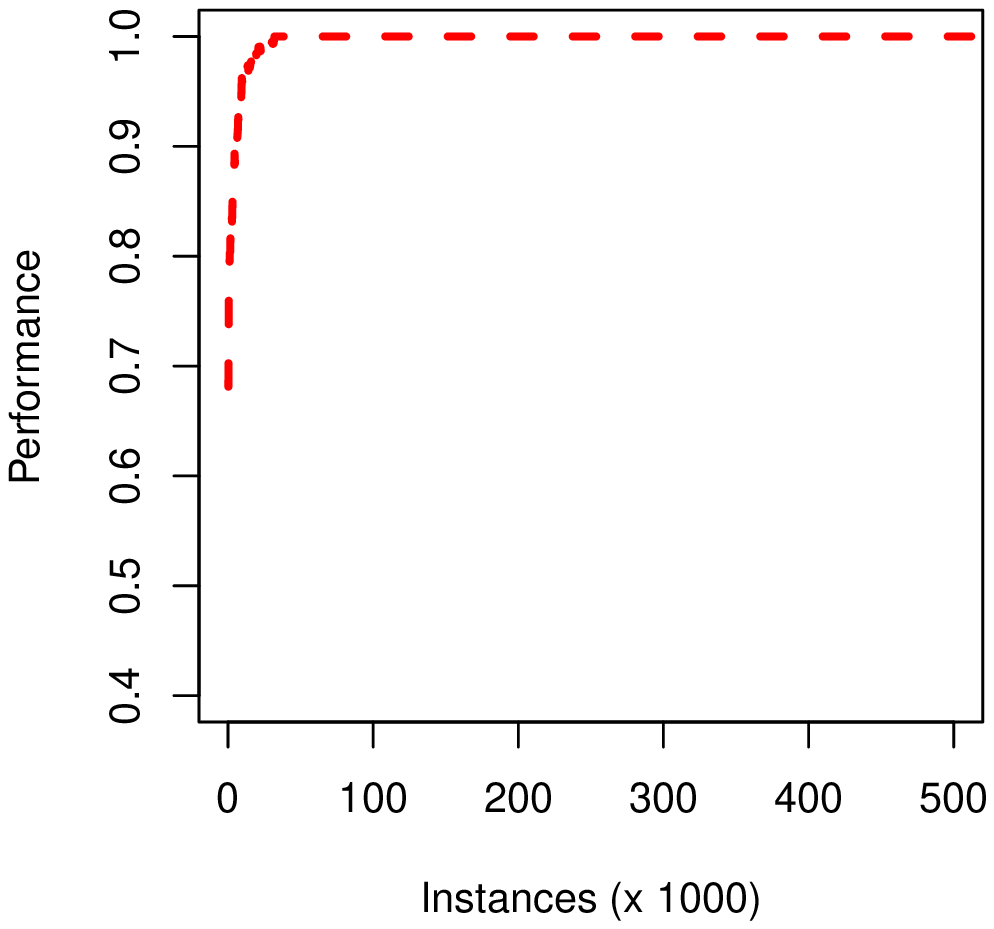}}	
	\captionsetup{justification=centering}	
	\caption{Learning curves of the subproblems of the Carry-one domain.}
	
	\label{fig:car_series}				
\end{figure*}
\begin{figure*}[!hbt]
	\centering
	\hspace{-0.40in}	
	\subfloat[SumModulo2]{
		\label{fig:par_series:par_0}		
		\includegraphics[width=1.81in]{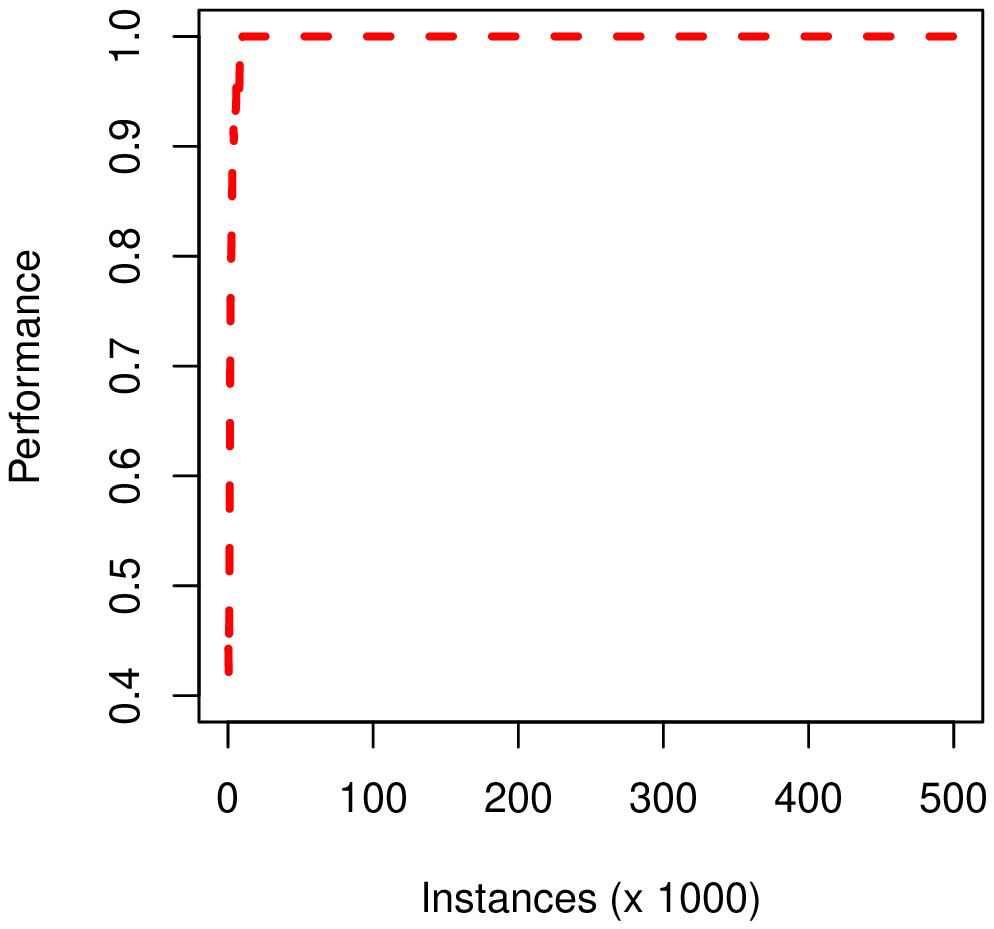}}
	\hspace{-0.01in}	
	\subfloat[isEvenParity]{
		\label{fig:par_series:par_1}		
		\includegraphics[width=1.81in]{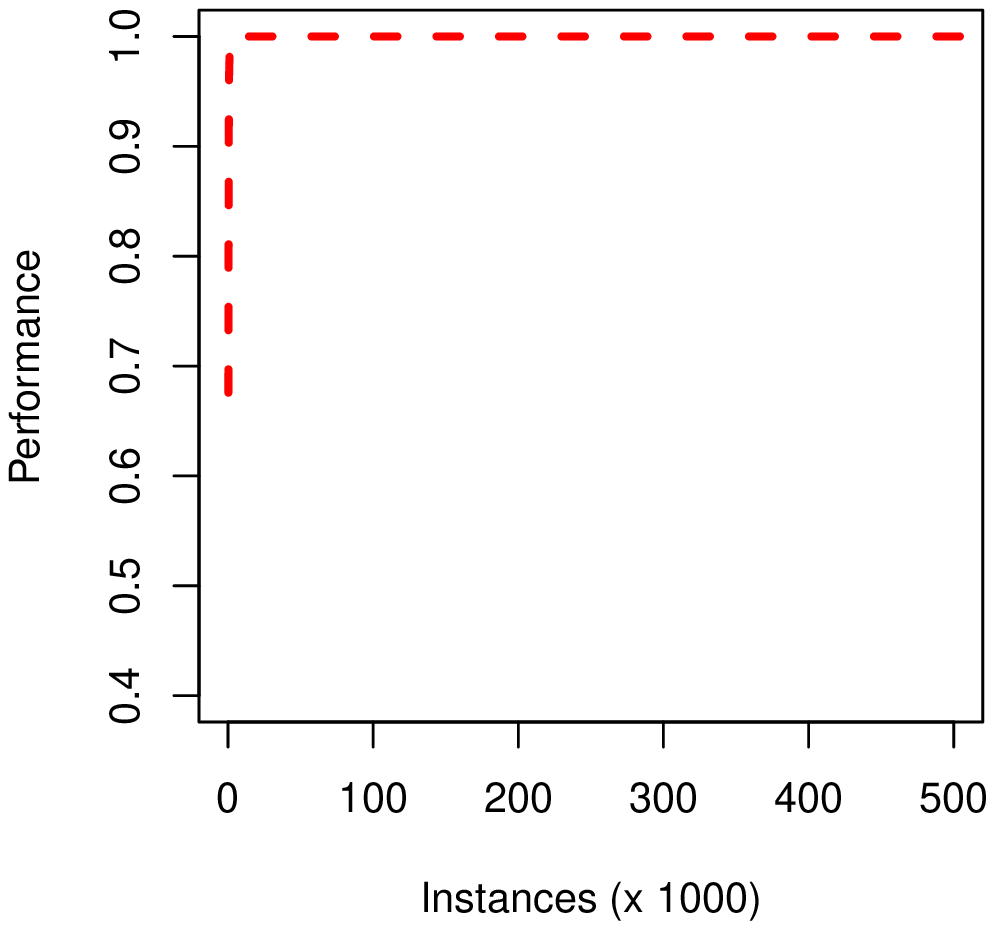}}	
	\hspace{-0.01in}
	\subfloat[isMajorityOn]{
		\label{fig:maj_series:maj_1}		
		\includegraphics[width=1.81in]{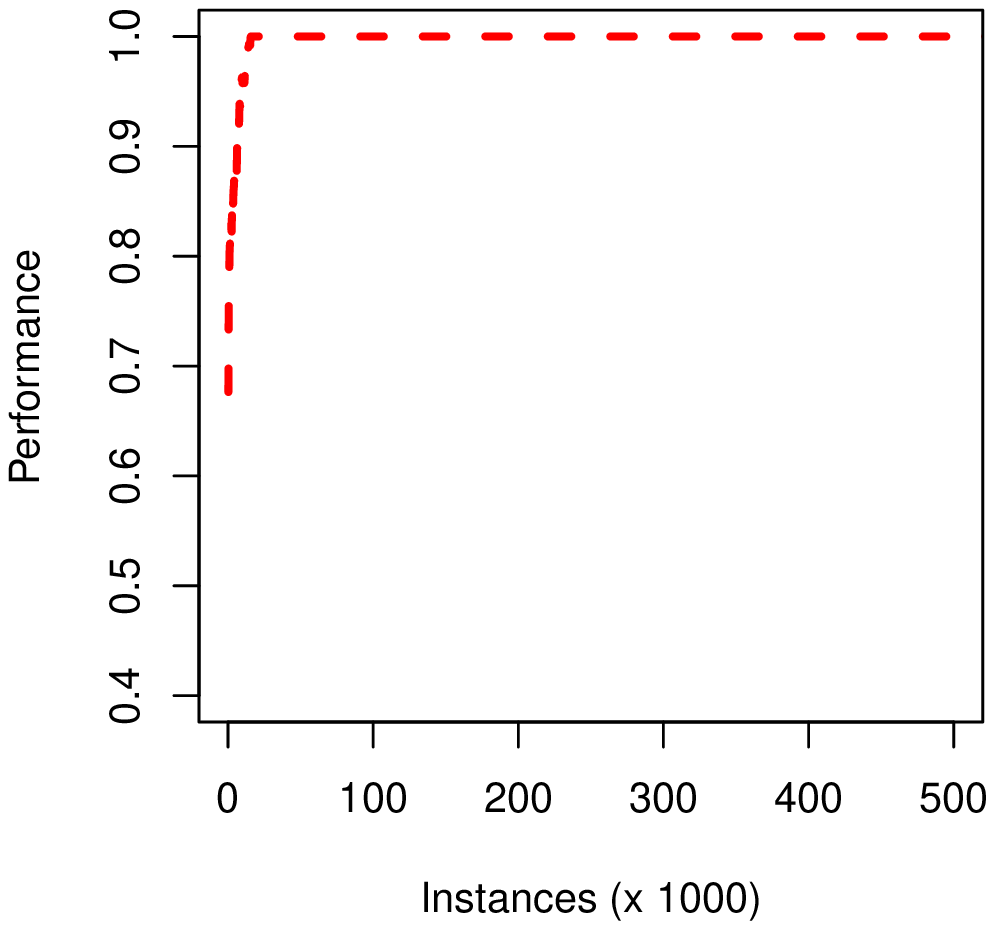}}	
	\captionsetup{justification=centering}
	\caption{Learning curves of the subproblems of the Even-parity ((a) and (b)) and Majority-on domains (c). The Majority-on domain also utilises the ``Half Length" subproblem of the Carry-one domain (\ref{fig:car_series:car_0}).}
	
	\label{fig:par_maj_series}				
\end{figure*}

Figures \ref{fig:mux_seris:mux_0} - \ref{fig:mux_seris:mux_5} show that training was successful in the sub-problems, which enabled XCSCF* to reuse the learned CF functionality of the Multiplexer problem. XCSCF* also successfully solved the subproblems of the Carry-one domain (Figures \ref{fig:car_series:car_0} - \ref{fig:car_series:car_5}), the Even-parity domain (Figures \ref{fig:par_series:par_0} and \ref{fig:par_series:par_1}), and the Majority-on domain (Figures \ref{fig:car_series:car_0} and \ref{fig:maj_series:maj_1}) (note the use of the HalfLength problem twice). The numbers of rules after compaction and CFs generated by all problems were generally only $1$, except for the ``Is Even-parity" problem with a little diversity of the genotypes of final solutions (see Section \ref{ssect:rules}). Reusing solutions from small-scale problems to solve large scale problems is plausible because maximally general rules are kept general without specific condition bits when used in larger-scale problems. This requires the logic behind the rule actions of the final solutions to be generalisable to the learned problems.

\begin{figure}
	\vspace{-3mm}
	\centering
	\includegraphics[width=3.2in]{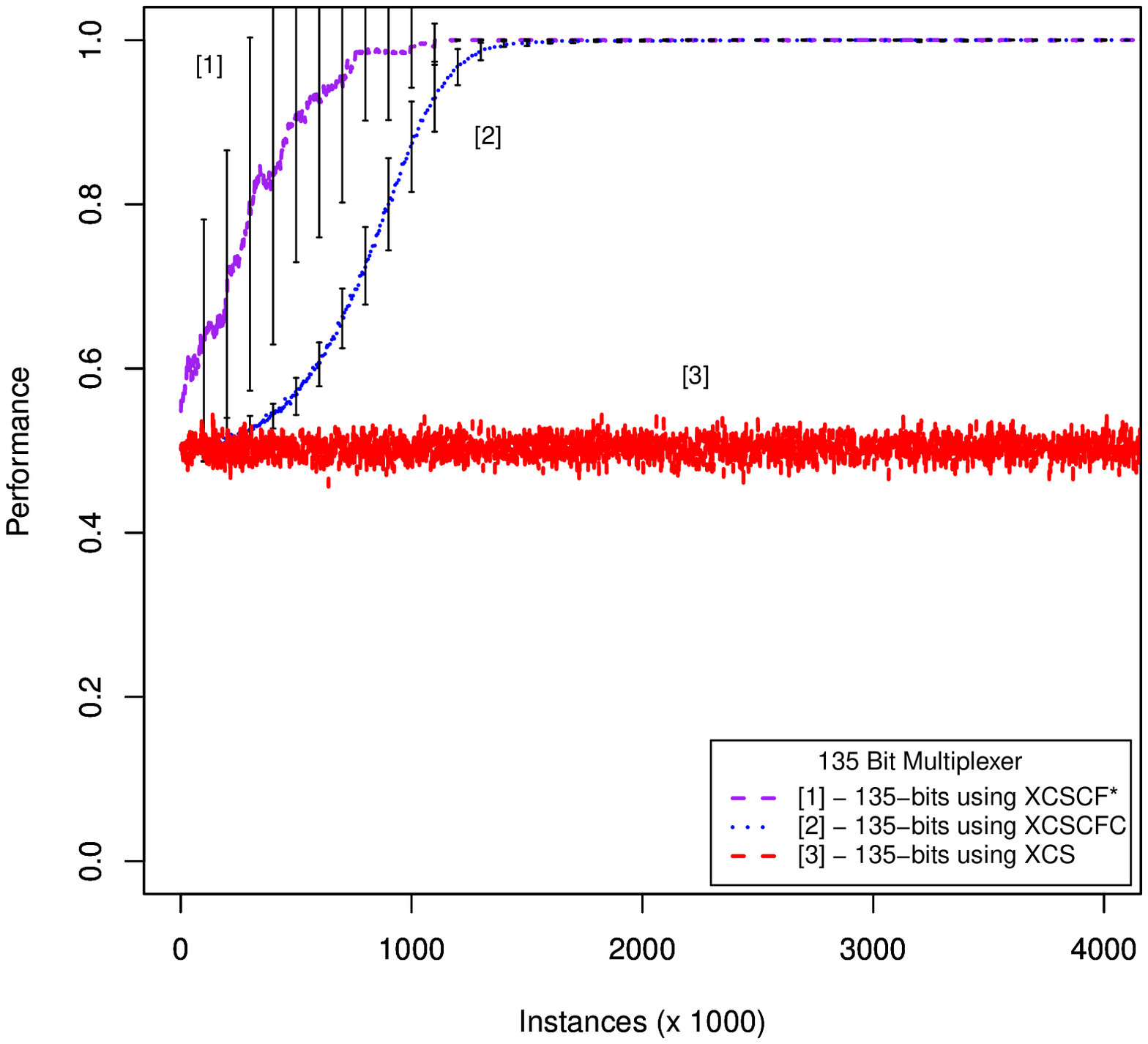}
	\caption{Performance of XCSCF* and XCSCFC on the 135-bit Multiplexer problem. Wilcoxon signed rank test shows no significant difference when converged.}
	\label{fig:135_Mux_Graph}
	\vspace{-1mm}
\end{figure}

Figure \ref{fig:135_Mux_Graph} shows that only the proposed system XCSCF* and XCSCFC were able to solve the 135-bit Multiplexer problem.  These experiments followed the standard explore and exploit phases of XCS. This shows scaling by relearning, but it is the capturing of the underlying patterns without retraining, which is the aim of this work.

Tests were conducted on the final rules produced by the final subproblem of the Multiplexer, the Carry-one, the Even-parity, and the Majority-on domains to determine if they were general enough to solve the corresponding problems at very large scales. Table \ref{tab:tests_large_scales} shows that the rule produced by the small-scale Multiplexer problem was able to solve the 1034-bit and even the 8205-bit Multiplexer problems \footnote{Note that $2^{8205}$ is a vast number, meaning that testing a million instances is a fractionally small sub-sample, but will identify many deficiencies.}. Similarly, the final rules of the final subproblem of other domains also achieved $100\%$ accuracies on corresponding problems at all tested large scales. The system used to test the generality of the rules was in straight exploitation: there was no covering, rule generation, or rule update. 

%
%

\begin{table}[!ht]
	\centering
	\caption{Accuracy tested on large-scale problems reusing solutions from final subproblems without training.}
	\begin{tabular}{|l|c|} 
		\hline
		Problems & Accuracies\\ \hline
		1034-bit Multiplexer & $100\%$ \\ \hline
		8205-bit Multiplexer & $100\%$ \\ \hline
		100-bit Carry-one & $100\%$ \\ \hline
		200-bit Carry-one & $100\%$ \\ \hline
		50-bit Even-parity & $100\%$ \\ \hline
		100-bit Even-parity & $100\%$ \\ \hline
		50-bit Majority-on & $100\%$ \\ \hline
		105-bit Majority-on & $100\%$ \\ \hline
	\end{tabular}
	\label{tab:tests_large_scales}
	\vspace{-3mm}
\end{table}

\subsection{Rules Generated by the Final Subproblems}\label{ssect:rules}

\begin{figure}[!ht]
	\vspace{-3mm}
	\centering
	\includegraphics[width=3.1in]{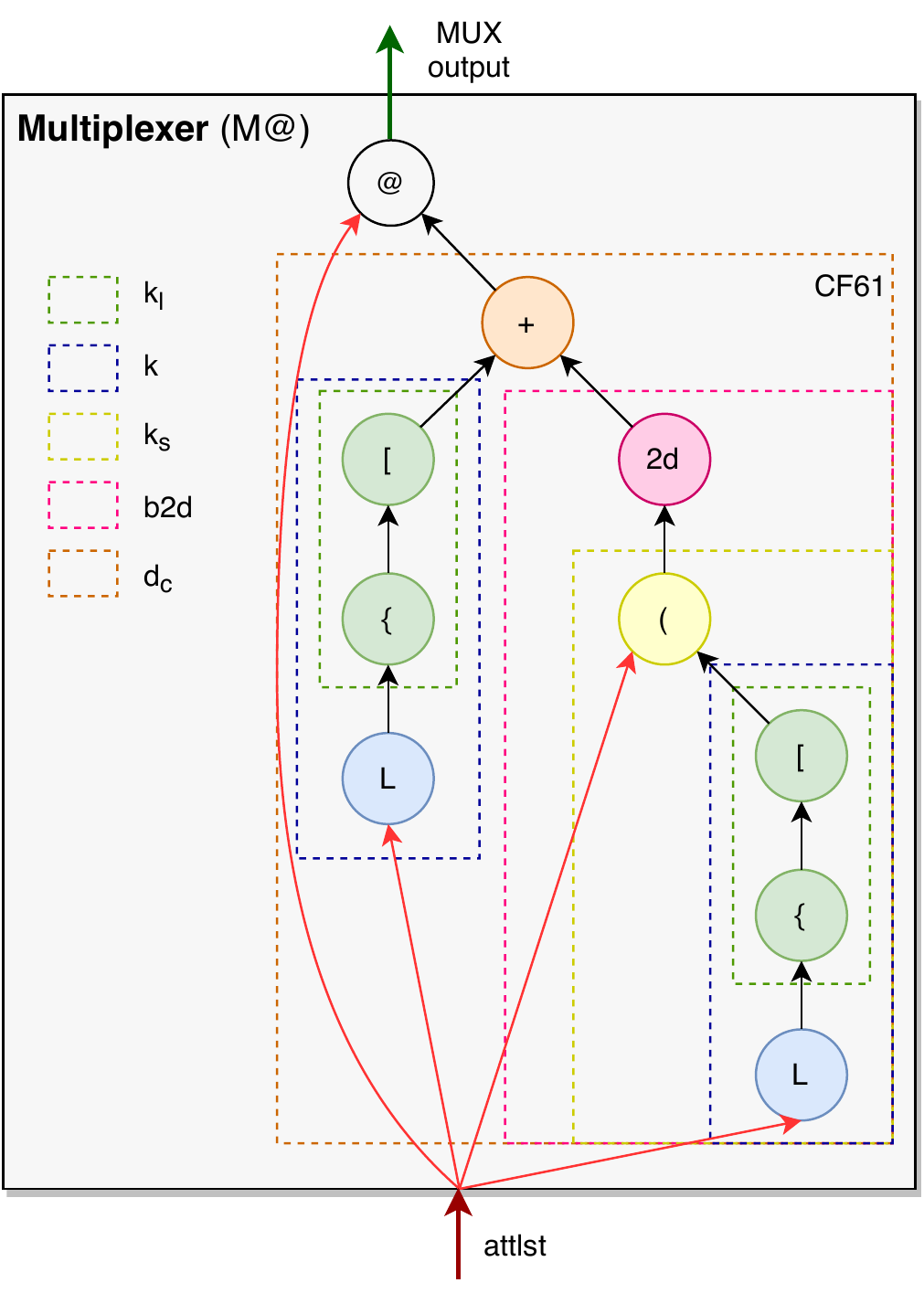}
	\caption{Multiplexer solution. Function nodes follow the tags in Table \ref{tab:skills_provided}. This solution uses nested learned functionalities in dashed boxes, which follow the tags in Table \ref{tab:functions_learned}.}
	\label{fig:Mux_Solution_Tree}
\end{figure}

The rule produced by the ``Multiplexer Data Bit" problem, the final subproblem of the Multiplexer domain, is illustrated in Table \ref{tab:6-bit_mux_rules_learned}. This rule is maximally general with no specified bit in its rule condition. Also, it seems very simple and neat for the general logic of the Multiplexer domain. This is not surprising given the functions accumulated by the experiential toolbox. The fully expanded tree in the rule action produced by the ``Multiplexer Data Bit" problem, the final subproblem of the Multiplexer domain, is illustrated in Figure \ref{fig:Mux_Solution_Tree}. Function nodes follow the function tags in Table \ref{tab:skills_provided}. The dashed boxes in this figure are the reused learned functions with names defined in Figure \ref{tab:functions_learned}. 

The tree in Figure \ref{fig:Mux_Solution_Tree} is the rule action of the one compacted rule for the n-bit Multiplexer problem. It accumulates a high-level function with many nested-layers of complexity. This complex tree can encapsulate the logic behind the n-bit Multiplexer problem through the guidance of all Multiplexer subproblems. For instance, the main building block of this tree is in the code fragment $CF61$, which provides the data bit position in the input bitstring using the function $d_c$ learned from the ``Data Bit Position" problem. This function $d_c$ is also a complex function involving an addition $(+)$ of the outputs from two reused functions within it, $k$ from the ``Multiplexer Address Length" problem and $b2d$ from the ``Multiplexer Data Channel" problem. The function $b2d$ converts the binary-string output of the function $k_s$ from the ``Multiplexer Address Bits" problem to a decimal value. $k_s$ returns the first ``Multiplexer Address Length" bits from the $attlst$ (the input bitstring) using the function $k$. The function $k$ is also nested function reusing a simpler function $k_l$ from the ``Multiplexer Address Length" problem (with Multiplexer scale as the input). The block $k$ is reused twice in the final solution $M@$ for the Multiplexer domain. The logic of the n-bit Multiplexer problem in the compacted rule with $M@$ in Table \ref{tab:6-bit_mux_rules_learned} was validated on the 1034-bit and 8205-bit Multiplexer problems (see Table \ref{tab:tests_large_scales}).

\begin{table}[ht]
	\centering
	\caption{Final rules learned before compaction while solving the ``Multiplexer Data Bit" problem, the final subproblem of the Multiplexer domain, cf. Figure \ref{fig:Mux_Solution_Tree}.}
	\begin{tabular}{l c} 
		\toprule
		Condition & Action \\ 
		\midrule
		\texttt{\# \# ... \# \#}  & \texttt{$attlst$ $CF61$ @}   \\ \bottomrule
	\end{tabular}
	\label{tab:6-bit_mux_rules_learned}
	\vspace{-3mm}
\end{table}

\begin{figure}[!ht]
	\centering
	\includegraphics[width=3.2in]{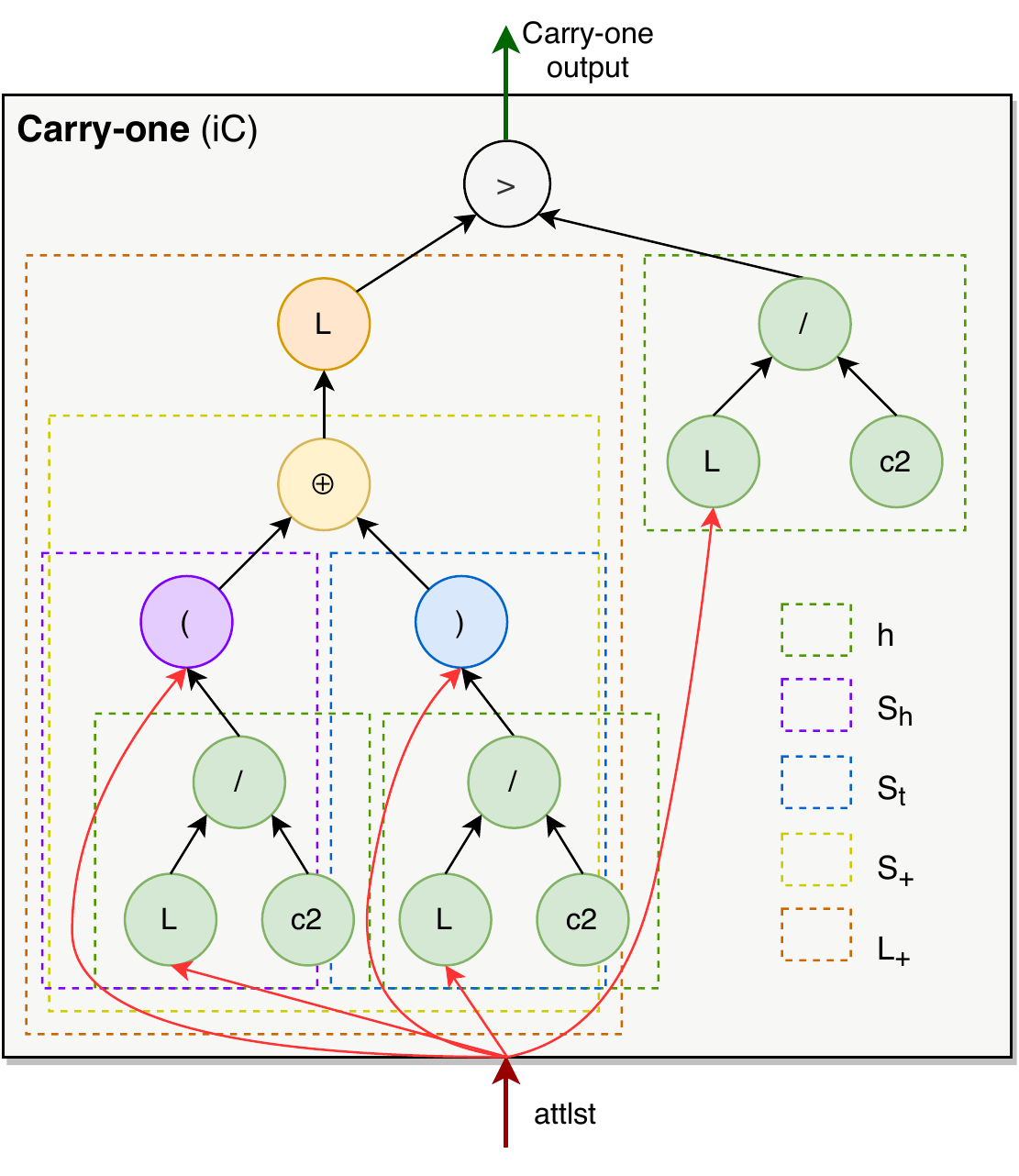}
	\caption{Carry-one solution.}
	\label{fig:Carry_Solution_Tree}
\end{figure}

\begin{figure}[!ht]
	\centering
	\includegraphics[width=3in]{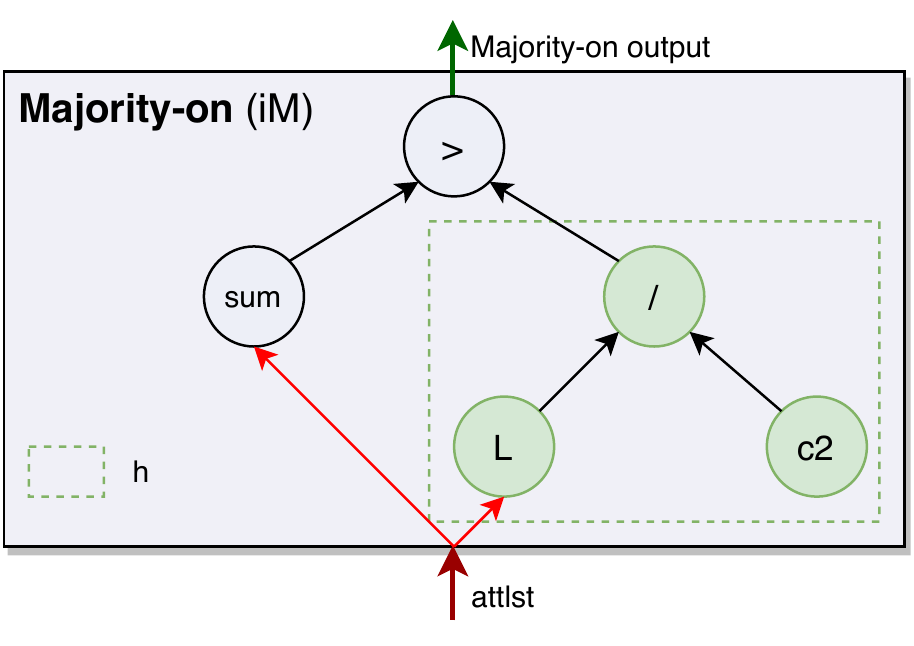}
	\caption{Majority-on solution.}
	\label{fig:majority_Solution_Tree}
	\vspace{-3mm}
\end{figure}

\begin{figure}[!ht]
	\centering
	\subfloat[Solution 1]{
		\label{fig:epar_solution:0}		
		\includegraphics[width=2.4in]{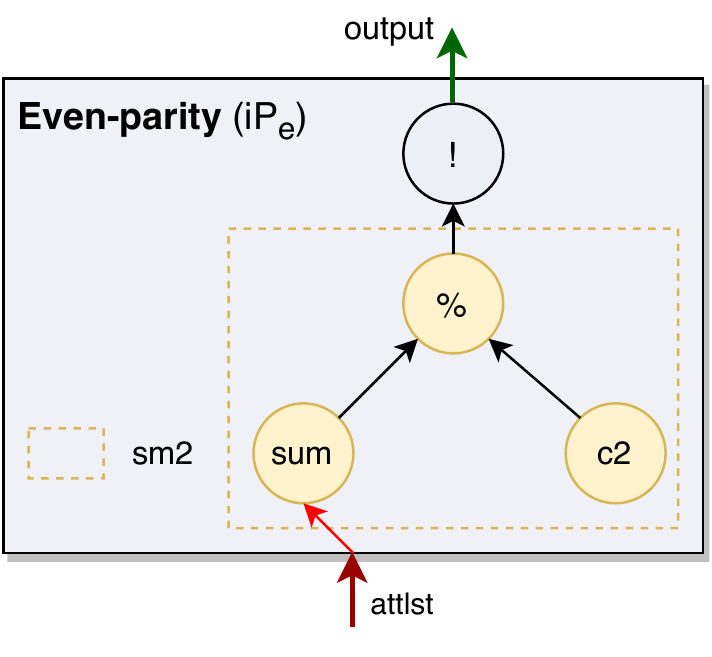}}
	\subfloat[Solution 2]{
		\label{fig:epar_solution:1}		
		\includegraphics[width=2.49in]{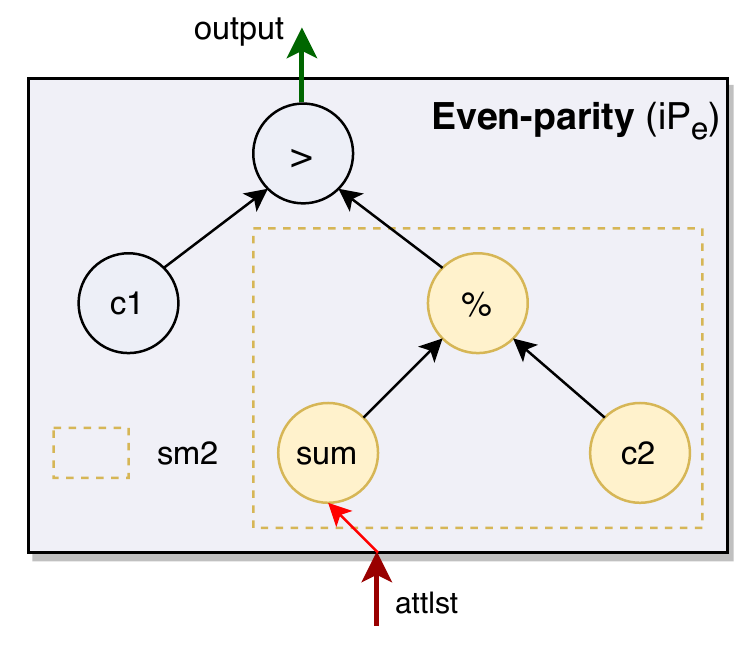}}
	\caption{Two most common solutions of the Even-parity domain. Solution 1 is more popularly discovered than solution 2.}
	\label{fig:epar_Solution_Tree}
	\vspace{-3mm}
\end{figure}

Other final rules of the Carry-one, Majority-on, and Even-parity domains also achieve maximal generality with all ``don't-care" bits in the condition part. These rules were also validated on the corresponding domains at very large scales. The trees in the rule actions of these final rules are illustrated respectively in Figure \ref{fig:Carry_Solution_Tree}, \ref{fig:majority_Solution_Tree}, and \ref{fig:epar_Solution_Tree}. Besides the Multiplexer domain, the Carry-one problem domain requires six subproblems to obtain the final logic, which resulted in a high complexity of the rule action. The final function $iC$ has five distinct nested functions within it and three occurrences of function $h$. The complexity of the solution for the n-bit Carry-one problem is equivalent to the complexity of function for the n-bit Multiplexer.

As the training flows of the Majority-on and Even-parity domains are straightforward, XCSCF* also discovered simpler rule actions in the final rules. XCSCF* yielded several different solutions for the Even-parity domains. The two most popular ones are illustrated in Figure \ref{fig:epar_Solution_Tree}. Solution 1 in this Figure appeared in most runs. Another solution found in only two runs is identical in logic to the solution 2, but the node $c1$ (constant CF of value $1$) is replaced with another CF that uses the division operator between $c1$ and a value of more than $1$.

\section{Discussions}

It can be said that the reason XCSCF* is capable of solving problems to a much larger scale than previously is that human knowledge separated the problem into appropriate and simpler sub-problems. Nevertheless, it is still a difficult task to learn each sub-task in such a manner that the learned knowledge/functionality could be transferred and then to learn to combine these blocks effectively. It is considered that the solutions of the tested problem domains, i.e. the Multiplexer, Carry-one, Even-parity, and Majority-on domains, yielded by XCSCF* contain the general logic of these domains and can solve these problems at any scale. 

The way that humans select sub-problems is similar to that of humans selecting function sets in standard EC approaches where too few or inappropriate selection prevents effective learning, while selecting too many unnecessary components could inhibit training. In these experiments, a number of redundant functions, such as the ceiling and the multiplication, and functions useful for only one specific problem domain, were never used by the final evolved solutions. XCSCF*, however, can identify the correct combination of accumulated knowledge to build complex solutions for the tested tasks.

Two main components of XCSCF* enables it to solve the tested problems fully. First, the supply of constants furnishes the required functionalities in the Carry-one, Even-parity, and Majority-on domains, as shown in the CFs of the final solutions. Second, the availability of the CF $attlst$ also contributes to solving the Carry-one, Majority-on, and Even-parity domains because it provides appropriate input for general functions, e.g. $StringSum$, $HeadList$, and $TailList$. It is argued that we can still input the environment state implicitly to all such functions. However, this method creates the complication of defining the environment state when these functions are nested in rule-set functions. Furthermore, deciding which functions should take the environment state by default and which functions should choose other string inputs requires extra human intervention. An extra benefit of using $attlst$ is that XCSCF* can now solve variable-scale problems in the tested domain. Previously, supplying a constant $L$ meant that the problem scale could not change.

It is evident that the proposed work has benefited from the transfer of learned information from each of the sub-problems. Reusing functionalities enables the system to achieve neat and abstract solutions although these solutions are actually complex without bloat when fully expanded. Although a defined recipe was not furnished to the system, it was able to form logical determinations as to the flow of the accumulated functionality, see Figure \ref{fig:Mux_Solution_Tree}.  This property of the system is similar to deriving a set of Threshold Concepts where significant learning towards the final target problem only advances once the proper chain of functionality is formed and evaluated.

\section{Conclusions and Future Works}

In this paper, we have introduced a developed LL system, i.e. XCSCF*, that can transfer learned knowledge and functionality. Starting from having minimal general knowledge (functions and skills) on the Boolean domain and some specific basic knowledge necessary for the target problems, XCSCF* is capable of learning general solutions to complex problem domains, i.e. the Multiplexer, Carry-one, Majority-on, and Even-parity domains, through analogies to the LL approach. By breaking down the problem domain into component sub-problems, providing the necessary axioms and transferring learned functionality in addition to knowledge, it is possible to identify general rules that can then be applied to any-scale problems in the domain. Another important observation is that not all of the provided functionality was utilised in the final solutions. 

Certain improvements of XCSCF* have been developed to enable learning the logic behind more general problems, such as the Multiplexer, Carry-one, Majority-on, and Even-parity domains. Removing the implicit connections between the instance and a few functions requires explicit connections between such functions and a newly provided attribute list, a more general input to replace the human-generated constant $L$. This explicit connections allows these functions to take any string-type inputs. Therefore, new XCSCF* provides more flexible logic and reduces the need for customisation to a given task. Also, the type-fitting property assures that generated CFs are compatible within themselves as well as with the target problem, which results in the ability to divide the search space by input and output types of available functions. Thus, this style of learning system can have access to more functionality than necessary for a single problem, without inhibited learning.

The general solutions from XCSCF* was validated by solving very difficult problems like the n-bit Multiplexer, n-bit Carry-one, n-bit Even-parity, and n-bit Majority-on problems. Although the aforementioned problems are comprised of a vastly sized search space, the proposed technique successfully discovered a minimal number (mostly one) of general rules. An advancement of this work was that the logic of complex problems was captured by simple trees when being described by the learned functionalities. However, once fully expanded, the CF trees contain certain complex nested patterns. Thus, LL can facilitate interpreting complex tree solutions using the intermediate learned components from the intermediate stages of LL.

Future work seeks to create a continuous-learning system with base axioms and a number of problems, including their possible subproblems, to be solved in a parallel architecture simultaneously. The \enquote*{toolbox} of functions (learned functions and axioms) plus the complementary knowledge (CFs) will grow as problems are solved and will be available for addressing future problems. The linked knowledge of solved problems would demonstrate interesting meta-knowledge, a form of learning curricula, and possible relationships among known problems, such as n-bit Multiplexer, n-bit Carry-one, etc. Furthermore, the research question is whether an XCS-based system with LL or parallel learning can solve real-valued datasets. The first thing to consider is to establish real-valued datasets that furnish LL.

\small

\bibliographystyle{apalike} 
\bibliography{solving_kbit_mux}

\end{document}


\maketitle
	
	
%
%
%
%
%
%
	

This supplementary material provides the samples of the training data for the problems described in Section 3.4 - Individual Detailed Components. The samples are shown in Table \ref{tab:kbitslength_training_data} to \ref{tab:isMajority}. The sample of each problem is grouped in a corresponding table. Note: as each dataset is used to form knowledge (if-then rules) plus the related function, their composition is crucial to overall/final learning of the system.
	

\begin{table}[ht]
	\centering
	\caption{Training data for the kBitsGivenLength sub-problem. It offers a mapping from a possible multiplexer length to the corresponding number of address bits. Shown are all eleven data points in this training set.}
	\begin{tabular}{r c} 
		\toprule
		Length &  Address Bits \\ 
		\midrule
		\texttt{3} & \texttt{1}   \\ \hline
		\texttt{6} & \texttt{2}   \\ \hline
		\texttt{11}  & \texttt{3}   \\ \hline
		\texttt{20}  & \texttt{4}   \\ \hline
		\texttt{37}  & \texttt{5}   \\ \hline
		\texttt{70}  & \texttt{6}   \\ \hline
		\texttt{135}  & \texttt{7}   \\ \hline
		\texttt{264}  & \texttt{8}   \\ \hline
		\texttt{521}  & \texttt{9}   \\ \hline
		\texttt{1034}  & \texttt{10}   \\ \hline
		\texttt{2059}  & \texttt{11}   \\ \bottomrule
	\end{tabular}
	\label{tab:kbitslength_training_data}
\end{table}


\begin{table}[ht]
	\centering
	\caption{Training data for the KBits sub-problem. It offers a mapping from the input bitstring of various Multiplexer problem scales to the corresponding number of address bits. This shows a sample of the $9$ data points. Note: the length of input varies in different scales, from $3$ to $37$.}
	\begin{tabular}{r c} 
		\toprule
		Bitstring &  Address Length \\ 
		\midrule
		\texttt{0 0 0} & \texttt{1}   \\ \hline
		\texttt{0 0 1} & \texttt{1}   \\ \hline
		\texttt{0 1 0} & \texttt{1}   \\ \hline
		\texttt{...} & \texttt{1}  \\ \hline
		\texttt{0 0 0 0 0 0} & \texttt{2}   \\ \hline
		\texttt{0 0 0 0 0 1}  & \texttt{2}   \\ \hline
		\texttt{0 0 0 0 1 0}  & \texttt{2}   \\ \hline
		\texttt{...} & \texttt{2/3/4}  \\ \hline
		\texttt{0 0 ... 0 0 (37 bits)}  & \texttt{5}   \\ \hline
		\texttt{0 0 ... 0 1 (37 bits)}  & \texttt{5}   \\ \hline
		\texttt{0 0 ... 1 0 (37 bits)}  & \texttt{5}   \\ \hline
		\texttt{...} & \texttt{5}  \\ \hline
	\end{tabular}
	\label{tab:kbits_training_data}
\end{table}


\begin{table}[ht]
	\centering
	\caption{A sample of the training data for the KBitString sub-problem with $8$ data points. This problem provides a mapping from the Multiplexer bitstrings to the corresponding lengths of the address bits. The scales range from $3$ to $20$ bits.}
	\begin{tabular}{r c} 
		\toprule
		Bitstring &  Address Length \\ 
		\midrule
		\texttt{0 0 0} & \texttt{0}   \\ \hline
		\texttt{...} & \texttt{...}  \\ \hline
		\texttt{1 1 1} & \texttt{1}   \\ \hline
		\texttt{0 0 0 0 0 0} & \texttt{0 0}   \\ \hline
		\texttt{...} & \texttt{...}  \\ \hline
		\texttt{0 1 0 0 0 0} & \texttt{0 1}   \\ \hline
		\texttt{...} & \texttt{...}  \\ \hline
		\texttt{1 0 0 0 0 0} & \texttt{0 1}   \\ \hline
		\texttt{...} & \texttt{...}  \\ \hline
		\texttt{1 1 1 1 1 1}  & \texttt{1 1}   \\ \hline
		\texttt{...} & \texttt{...}  \\ \hline
		\texttt{0 0 0 0 ... 0 0 (20 bits)}  & \texttt{0 0 0 0}   \\ \hline
		\texttt{...}  & \texttt{...}   \\ \hline
		\texttt{1 1 1 1 ... 1 1 (20 bits)}  & \texttt{1 1 1 1}   \\ \hline
	\end{tabular}
	\label{tab:kbit_string_training_data}
\end{table}


\begin{table}[ht]
	\centering
	\caption{A sample of the training data for the Bin2Int sub-problem with $8$ instances. The expected data channel index is the position of the Multiplexer data channel (among data channels only). The scales vary from $3$ to $20$ bits.}
	\begin{tabular}{r c}
		\toprule
		Bitstring &  Data Channel index \\ 
		\midrule
		\texttt{0 0 0} & \texttt{0}   \\ \hline
		\texttt{...} & \texttt{...}  \\ \hline
		\texttt{1 1 1} & \texttt{1}   \\ \hline
		\texttt{0 0 0 0 0 0} & \texttt{0}   \\ \hline
		\texttt{...} & \texttt{...}  \\ \hline
		\texttt{0 1 0 0 0 0} & \texttt{1}   \\ \hline
		\texttt{...} & \texttt{...}  \\ \hline
		\texttt{1 0 0 0 0 0} & \texttt{2}   \\ \hline
		\texttt{...} & \texttt{...}  \\ \hline
		\texttt{1 1 1 1 1 1}  & \texttt{3}   \\ \hline
		\texttt{...} & \texttt{...}  \\ \hline
		\texttt{0 0 0 0 ... 0 0 (20 bits)}  & \texttt{0}   \\ \hline
		\texttt{...}  & \texttt{...}   \\ \hline
		\texttt{1 1 1 1 ... 1 1 (20 bits)}  & \texttt{15}   \\ \hline
	\end{tabular}
	\label{tab:bin_to_int_training_data}
\end{table}


\begin{table}[ht]
	\centering
	\caption{A sample with $8$ data points for the training data for the AddressOf sub-problem. The expect output is the position of the Multiplexer data channel among all bits in the Multiplexer input bitstring. The scales of this problem ranges from $3$ to $20$ bits.}
	\begin{tabular}{r c} 
		\toprule
		Bitstring & \specialcell{Attribute position of\\the Data Channel} \\ 
		\midrule
		\texttt{0 0 0} & \texttt{1}   \\ \hline
		\texttt{...} & \texttt{...}  \\ \hline
		\texttt{1 1 1} & \texttt{2}   \\ \hline
		\texttt{0 0 0 0 0 0} & \texttt{2}   \\ \hline
		\texttt{...} & \texttt{...}  \\ \hline
		\texttt{0 1 0 0 0 0} & \texttt{3}   \\ \hline
		\texttt{...} & \texttt{...}  \\ \hline
		\texttt{1 0 0 0 0 0} & \texttt{4}   \\ \hline
		\texttt{...} & \texttt{...}  \\ \hline
		\texttt{1 1 1 1 1 1}  & \texttt{5}   \\ \hline
		\texttt{...} & \texttt{...}  \\ \hline
		\texttt{0 0 0 0 ... 0 0 (20 bits)}  & \texttt{4}   \\ \hline
		\texttt{...}  & \texttt{...}   \\ \hline
		\texttt{1 1 1 1 ... 1 1 (20 bits)}  & \texttt{19}   \\ \hline
	\end{tabular}
	\label{tab:address_of_training_data}
\end{table}


\begin{table}[ht]
	\centering
	\caption{A sample of $8$ data points for the training data for the ValueAt sub-problem. The bold bit in the bitstring is the channel connecting to the output of the Multiplexer and also the expected output of this subproblem. The scales vary from $3$ to $20$ bits.}
	\begin{tabular}{r c} 
		\toprule
		Bitstring & \specialcell{Attribute position of\\the Data Channel} \\ 
		\midrule
		{0 \textbf{0} 0} & \texttt{0}   \\ \hline
		{...} & \texttt{...}  \\ \hline
		{1 1 \bf1} & \texttt{1}   \\ \hline
		{0 0 \textbf{0} 0 0 0} & \texttt{0}   \\ \hline
		{...} & \texttt{...}  \\ \hline
		{0 1 0 \textbf{0} 0 0} & \texttt{0}   \\ \hline
		{...} & \texttt{...}  \\ \hline
		{1 0 0 0 \textbf{0} 0} & \texttt{0}   \\ \hline
		{...} & \texttt{...}  \\ \hline
		{1 1 1 1 1 \bf1}  & \texttt{1}   \\ \hline
		{...} & \texttt{...}  \\ \hline
		{0 0 0 0 \bf0 ... 0 0 (20 bits)}  & \texttt{0}   \\ \hline
		{...}  & \texttt{...}   \\ \hline
		{1 1 1 1 1 ... 1 \textbf{1} (20 bits)}  & \texttt{1}   \\ \hline
	\end{tabular}
	\label{tab:value_at_training_data}
\end{table}


\begin{table}[ht]
	\centering
	\caption{$6$ samples of the training data for the SumMod2 sub-problem. The expected output is the summation of the input bitstring modulo $2$. The problem scales ranges from $1$ to $8$ bits.}
	\begin{tabular}{c c} 
		\toprule
		BitString & Output  \\ 
		\midrule
		\texttt{0} & \texttt{0}   \\ \hline
		\texttt{1} & \texttt{1}   \\ \hline
		\texttt{0 0} & \texttt{0}   \\ \hline
		\texttt{0 1} & \texttt{1}   \\ \hline
		\texttt{...} & \texttt{0/1}           \\ \hline
		\texttt{1 1 1 1 1 1 1 0} & \texttt{1}    \\ \hline
		\texttt{1 1 1 1 1 1 1 1} & \texttt{0}    \\ \bottomrule
	\end{tabular}
	\label{tab:sum_mod2}
\end{table}


\begin{table}[ht]
	\centering
	\caption{$4$ samples of the training data for the isEvenParity sub-problem. The expected output of this problem is equivalent to the output of a Even-parity problem with varied scales, which returns $1$ when the summation of the input is even and $0$ otherwise. The problem scales ranges from $1$ to $11$ bits.}
	\begin{tabular}{c c} 
		\toprule
		BitString & Output  \\ 
		\midrule
		\texttt{0} & \texttt{1}   \\ \hline
		\texttt{1} & \texttt{0}   \\ \hline
		\texttt{0 0} & \texttt{1}   \\ \hline
		\texttt{0 1} & \texttt{0}   \\ \hline
		\texttt{...} & \texttt{0/1}           \\ \hline
		\texttt{1 1 1 1 1 1 1 1 1 1 0} & \texttt{1}    \\ \hline
		\texttt{1 1 1 1 1 1 1 1 1 1 1} & \texttt{0}    \\ \bottomrule
	\end{tabular}
	\label{tab:isEvenParity}
\end{table}


\begin{table}[ht]
	\centering
	\caption{$4$ samples of the training data for the HalfLength sub-problem. This problem expects the output to be half the length of the input bitstring. The scales are even numbers from $4$ to $16$.}
	\begin{tabular}{c c} 
		\toprule
		BitString & Output  \\ 
		\midrule
		\texttt{0 0 0 0} & \texttt{2}   \\ \hline
		\texttt{0 0 0 1} & \texttt{2}   \\ \hline
		\texttt{...} & \texttt{...}           \\ \hline
		\texttt{1 1 1 1 1 1 1 1 1 1 1 1 1 1 1 0} & \texttt{8}    \\ \hline
		\texttt{1 1 1 1 1 1 1 1 1 1 1 1 1 1 1 1} & \texttt{8}    \\ \bottomrule
	\end{tabular}
	\label{tab:HalfLength}
\end{table}


\begin{table}[ht]
	\centering
	\caption{A sample of $4$ instances of the training data for the HeadString sub-problem. The bold part in a bitstring correspond to the output, which is the first half of the input and is considered as a binary number. The scales are even numbers from $4$ to $10$.}
	\begin{tabular}{c c} 
		\toprule
		BitString & Output  \\ 
		\midrule
		{\textbf{0 0} 0 0} & \texttt{0 0}   \\ \hline
		{\textbf{0 0} 0 1} & \texttt{0 0}   \\ \hline
		\texttt{...} & \texttt{...}           \\ \hline
		{\textbf{1 1 1 1 1} 1 1 1 1 0} & \texttt{0 1 1 1 1}    \\ \hline
		{\textbf{1 1 1 1 1} 1 1 1 1 1} & \texttt{1 1 1 1 1}    \\ \bottomrule
	\end{tabular}
	\label{tab:HeadString}
\end{table}


\begin{table}[ht]
	\centering
	\caption{Training data for the TailString sub-problem. The bold part in a bitstring correspond to the output, which is the second half of the input. This output is also considered as a binary number. The scales are even numbers from $4$ to $10$.}
	\begin{tabular}{c c} 
		\toprule
		BitString & Output  \\ 
		\midrule
		{0 0 \textbf{0 0}} & \texttt{0 0}   \\ \hline
		{0 0 \textbf{0 1}} & \texttt{0 0}   \\ \hline
		\texttt{...} & \texttt{...}           \\ \hline
		{0 0 0 0 0 \textbf{0 0 0 0 0}} & \texttt{0 0 0 0 0}    \\ \hline
		{0 0 0 0 0 \textbf{0 0 0 0 1}} & \texttt{0 0 0 0 1}    \\ \hline
		\texttt{...} & \texttt{...}           \\ \bottomrule
	\end{tabular}
	\label{tab:TailString}
\end{table}


\begin{table}[ht]
	\centering
	\caption{$4$ samples of the training data for the BinarySum sub-problem. The output is the binary summation of the two binary numbers represented by the first half and the second half of the input bitstring.}
	\begin{tabular}{c c} 
		\toprule
		BitString & Output  \\ 
		\midrule
		\texttt{0 0 0 0} & \texttt{0 0}   \\ \hline
		\texttt{0 0 0 1} & \texttt{0 1}   \\ \hline
		\texttt{...} & \texttt{...}           \\ \hline
		\texttt{1 1 1 1 1 1 1 1 1 0} & \texttt{1 1 1 1 0 1}    \\ \hline
		\texttt{1 1 1 1 1 1 1 1 1 1} & \texttt{1 1 1 1 1 0}    \\ \bottomrule
	\end{tabular}
	\label{tab:BinarySum}
\end{table}


\begin{table}[ht]
	\centering
	\caption{$4$ samples of the training data for the SumStringLength sub-problem. The expected output is the length of the output of the BinarySum problem, which is the binary summation of the two binary numbers represented by the two halves of the input bitstring.}
	\begin{tabular}{c c}
		\toprule
		BitString & Output  \\ 
		\midrule
		\texttt{0 0 0 0} & \texttt{2}   \\ \hline
		\texttt{0 0 0 1} & \texttt{2}   \\ \hline
		\texttt{...} & \texttt{...}           \\ \hline
		\texttt{1 1 1 1 1 1 1 1 1 0} & \texttt{6}    \\ \hline
		\texttt{1 1 1 1 1 1 1 1 1 1} & \texttt{6}    \\ \bottomrule
	\end{tabular}
	\label{tab:SumStringLength}
\end{table}


\begin{table}[ht]
	\centering
	\caption{$4$ samples of the training data for the isCarried sub-problem. The output of this problem is $1$ when the binary summation of the two binary numbers (represented by the two halves of the input) carries $1$ in the highest bit, and $0$ otherwise. The problem scales are all even numbers from $4$ to $12$ bits.}
	\begin{tabular}{c c} 
		\toprule
		BitString & Output  \\ 
		\midrule
		\texttt{0 0 0 0} & \texttt{0}   \\ \hline
		\texttt{0 0 0 1} & \texttt{0}   \\ \hline
		\texttt{...} & \texttt{...}           \\ \hline
		\texttt{1 1 1 1 1 1 1 1 1 1 1 0} & \texttt{1}    \\ \hline
		\texttt{1 1 1 1 1 1 1 1 1 1 1 1} & \texttt{1}    \\ \bottomrule
	\end{tabular}
	\label{tab:isCarried}
\end{table}


\begin{table}[ht]
	\centering
	\caption{A sample of $12$ instances of the training data for the isMajorityOn sub-problem. The output is $1$ when the number of bits $1$ is greater than the half length of the input bitstring, and $0$ otherwise. The scales vary from $1$ to $7$ bits.}
	\begin{tabular}{c c} 
		\toprule
		BitString & Output  \\ 
		\midrule
		\texttt{0} & \texttt{0}   \\ \hline
		\texttt{1} & \texttt{1}   \\ \hline
		\texttt{0 0} & \texttt{0}   \\ \hline
		\texttt{0 1} & \texttt{0}   \\ \hline
		\texttt{1 0} & \texttt{0}   \\ \hline
		\texttt{1 1} & \texttt{1}   \\ \hline
		\texttt{...} & \texttt{...}           \\ \hline
		\texttt{0 0 0 0 0 0 0} & \texttt{0}    \\ \hline
		\texttt{...} & \texttt{...}           \\ \hline
		\texttt{0 0 0 1 1 1 0} & \texttt{0}    \\ \hline
		\texttt{0 0 0 1 1 1 1} & \texttt{1}    \\ \hline
		\texttt{0 0 1 0 0 0 0} & \texttt{0}    \\ \hline
		\texttt{...} & \texttt{...}           \\ \hline
		\texttt{1 1 1 1 1 1 0} & \texttt{1}    \\ \hline
		\texttt{1 1 1 1 1 1 1} & \texttt{1}    \\ \bottomrule
	\end{tabular}
	\label{tab:isMajority}
\end{table}
	